\documentclass[sigconf]{acmart}
\AtBeginDocument{%
  }

\copyrightyear{2026}
\acmYear{2026}
\setcopyright{cc}
\setcctype{by}
\acmConference[KDD '26]{Proceedings of the 32nd ACM SIGKDD Conference on Knowledge Discovery and Data Mining V.2}{August 09--13, 2026}{Jeju Island, Republic of Korea}
\acmBooktitle{Proceedings of the 32nd ACM SIGKDD Conference on Knowledge Discovery and Data Mining V.2 (KDD '26), August 09--13, 2026, Jeju Island, Republic of Korea}
\acmDOI{10.1145/3770855.3817808}
\acmISBN{979-8-4007-2259-2/2026/08}

\usepackage{amsmath,amsfonts,bm}

\def\eqref#1{equation~\ref{#1}}

\def\1{\bm{1}}

\def\vh{{\bm{h}}}

\def\vp{{\bm{p}}}

\def\vx{{\bm{x}}}

\def\mA{{\bm{A}}}
\def\mB{{\bm{B}}}
\def\mC{{\bm{C}}}

\def\mH{{\bm{H}}}
\def\mI{{\bm{I}}}
\def\mJ{{\bm{J}}}

\def\mP{{\bm{P}}}
\def\mQ{{\bm{Q}}}

\def\mU{{\bm{U}}}
\def\mV{{\bm{V}}}

\def\mX{{\bm{X}}}

\DeclareMathAlphabet{\mathsfit}{\encodingdefault}{\sfdefault}{m}{sl}
\SetMathAlphabet{\mathsfit}{bold}{\encodingdefault}{\sfdefault}{bx}{n}

\def\sR{{\mathbb{R}}}

\usepackage{multirow}
\usepackage{wrapfig}
\usepackage{xcolor}
\usepackage{balance}
\usepackage{listings}
\usepackage{colortbl}
\usepackage{subcaption}
\definecolor{highlightblue}{RGB}{175, 177, 214}

\begin{document}


\title{Native Hierarchical and Compositional Representations with Subspace Embeddings}

\author{Gabriel Moreira}
\orcid{0000-0001-8564-2835}
\affiliation{%
  \institution{ISR, Instituto Superior Técnico}
  \city{Lisbon}
  \country{Portugal}
}
\affiliation{%
  \institution{Carnegie Mellon University}
  \city{Pittsburgh}
  \state{PA}
  \country{USA}
}
\email{gmoreira@cs.cmu.edu}

\author{Zita Marinho}
\orcid{0000-0002-6314-6096}
\affiliation{%
  \institution{IT, ISR, Instituto Superior Técnico}
  \city{Lisbon}
  \country{Portugal}}
\email{zita.marinho@tecnico.ulisboa.pt}

\author{Manuel Marques}
\orcid{0000-0003-0532-1869}
\affiliation{%
  \institution{ISR, Instituto Superior Técnico}
  \city{Lisbon}
  \country{Portugal}}
\email{manuel@isr.tecnico.ulisboa.pt}

\author{João Paulo Costeira}
\orcid{0000-0001-6769-2935}
\affiliation{%
  \institution{ISR, Instituto Superior Técnico}
  \city{Lisbon}
  \country{Portugal}}
\email{jpc@isr.tecnico.ulisboa.pt}

\author{Chenyan Xiong}
\orcid{0000-0002-0392-4183}
\affiliation{%
  \institution{Carnegie Mellon University}
  \city{Pittsburgh}
  \state{PA}
  \country{USA}
}
\email{cx@cs.cmu.edu}


\begin{abstract}
    Traditional embeddings represent datapoints as vectors, which makes similarity easy to compute but limits how well they capture hierarchies and compositionality. We propose a fundamentally different approach: representing concepts as linear subspaces. By spanning multiple dimensions, subspaces can model broader concepts with higher-dimensional regions and nest more specific concepts within them. This geometry naturally captures generality through dimension, hierarchy through inclusion, and enables an emergent structure for composition via linear algebraic operations. To make this paradigm trainable, we introduce a differentiable subspace parameterization via soft projection matrices, allowing the effective dimension of each subspace to be learned. Our method not only achieves state-of-the-art performance on hierarchical and natural language inference benchmarks but also provides a geometrically-grounded model of entailment. Further, we demonstrate that while standard vector embeddings degrade to near-random performance on negated queries, subspace embeddings natively capture logical composition without explicit supervision, while preserving compatibility with efficient Euclidean vector search.
\end{abstract}




\maketitle
\newcommand\kddavailabilityurl{https://doi.org/10.5281/zenodo.20357423}
\ifdefempty{\kddavailabilityurl}{}{
\begingroup\small\noindent\raggedright\textbf{Resource Availability:}\\
The source code of this paper has been made publicly available at \url{\kddavailabilityurl}.
\endgroup
}

\section{Introduction}
\label{sec:intro}

Dense vector embeddings have become the bedrock of modern machine learning, underpinning systems from language models (LMs) \citep{devlin2019bert,reimers2019sentence} and vision-language models (VLMs) \citep{radford2021learning, li2022blip} to retrieval augmented generation (RAG) systems \citep{lewis2020retrieval}. By representing words, documents, and images as points in high-dimensional space, these representations excel at capturing similarities in a scalable manner. 

Despite their success, the efficacy of vector embeddings is limited by a geometric mismatch: the flat, symmetric structure of Euclidean space is ill-suited to the hierarchical and asymmetric nature of language and semantic reasoning. Due to their symmetry, metrics like cosine similarity cannot capture directional relationships such as entailment or hyponymy; a high similarity between ``dog'' and ``animal'' fails to convey that one subsumes the meaning of the other. Moreover, vector spaces lack native operators for conjunction and negation. This forces models to default to additive composition, effectively treating phrases as a bag of words. This explains why queries with negations often fail, with embeddings retaining the very concept meant for exclusion. Recent work confirms these flaws empirically, showing that even state-of-the-art models disregard logical connectives \citep{yuksekgonul2022and, moreira2025learning}, requiring \textit{ad hoc} solutions \citep{weller2023nevir,gokhale2020vqa,zhang2025negvqavisionlanguagemodels,alhamoud2025vision}. This inability to interpret nuanced instructions motivates our search for a representation that natively encodes these relations.

\begin{figure*}[t]
    \centering
    \includegraphics[width=1.0\linewidth]{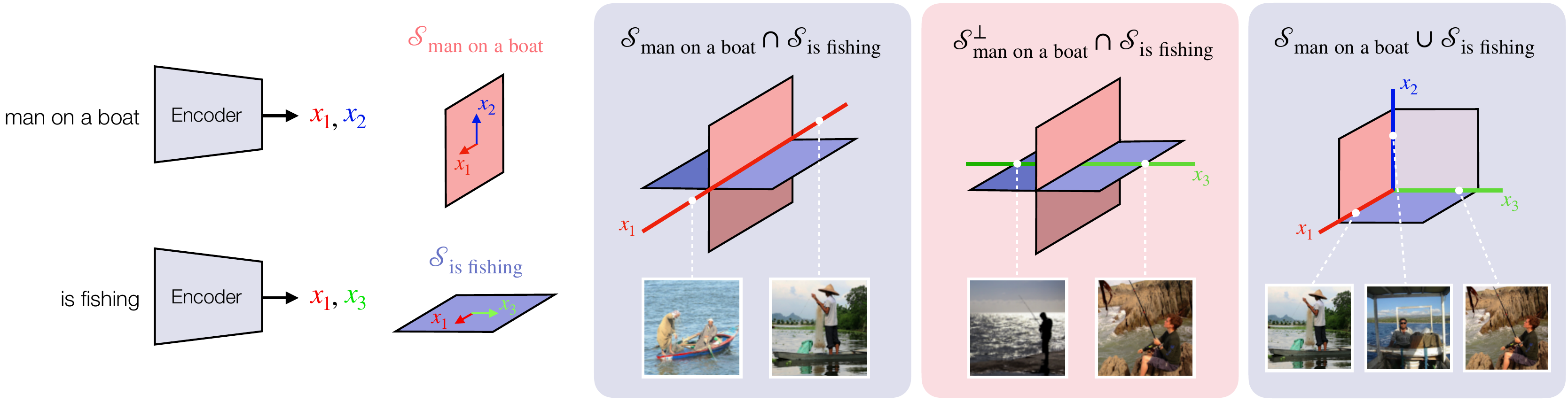}
    \caption{We embed concepts as linear subspaces of $\mathbb{R}^d$ (left). These representations enable logical operations: subspace intersections \textit{e.g.}, ``man on a boat'' and ``is fishing'' (middle left); negation and composition \textit{e.g.}, orthogonal complement of ``man on a boat'' and ``is fishing'' (middle right) and linear sums of subspaces, which yield a higher variance of instances (right).}
    \label{fig:overview}
\end{figure*}

We propose an alternative that extends Euclidean vector representations: instead of mapping a concept to a single vector, we embed it as a linear subspace \textit{i.e.}, the span of a set of basis vectors. This enables an interpretable geometric understanding of conceptual properties. First, generality and specificity are captured by the subspace dimension, with higher-dimensional subspaces denoting broader concepts \textit{e.g.}, animal vs. dog. Second, hierarchy is naturally modeled by subspace inclusion, where a more specific concept's subspace is contained within a more general one. Finally, logical operations are directly mapped to linear-algebraic operations: conjunction as subspace intersection, disjunction as linear sum (span), and negation as the orthogonal complement (Fig. \ref{fig:overview}).

A key challenge in learning subspaces is that dimensionality, or the number of basis vectors, is discrete and thus non-differentiable. Our technical contribution overcomes this by introducing a differentiable parameterization via soft projection matrices. Instead of selecting an integer dimension, we learn a set of vectors and modulate their individual importance, allowing each subspace to add or drop basis vectors as needed during training. Crucially, this approach remains grounded in Euclidean geometry, preserving full compatibility with standard training pipelines, Euclidean metrics and loss functions. This allows for seamless integration with efficient dot-product-based search libraries \cite{douze2024faiss, johnson2019billion}, ensuring scalability.

We validate our framework on hierarchical modeling as well as lexical and textual entailment benchmarks. Our method sets a new state-of-the-art on \textsc{WordNet} reconstruction, shows a stronger correlation with human judgments on HyperLex, and outperforms bi-encoder baselines on SNLI. Crucially, we demonstrate that standard vector baselines degrade to near-random performance on compositional entailment with negated hypotheses (49--69\% AUC). In contrast, subspace embeddings maintain discriminative power ($>$90\% AUC) without explicit supervision.

Beyond quantitative accuracy, our approach offers significant practical advantages. We observe a strong correlation between the learned dimensionality of a subspace and the semantic specificity of the concept it represents. This yields an interpretable measure of information content as an emergent property that arises from the geometry of the representations. Further, unlike hyperbolic embeddings, which rely on non-Euclidean metrics, our representations are compatible with standard Maximum Inner Product Search, enabling retrieval nearly $8\times$ faster than Poincaré baselines.

In summary, our key contributions are:
\begin{itemize}
\item A novel differentiable method for learning subspace representations of language, where data-dependent dimensionality captures semantic specificity.
\item An emergent structure for composition over natural language. We show that logical operations, such as conjunction, disjunction, and negation, arise naturally from standard training objectives, without any explicit supervision.
\item Empirical evidence that subspace embeddings remain tractable for large-scale retrieval by preserving compatibility with Euclidean vector search pipelines.
\end{itemize}

\section{Related Work}
\label{sec:related_work}

Vector embeddings, from Word2Vec \citep{mikolov2013distributed} to
dense retrievers \citep{reimers2019sentence,wang2024multilingual}
and LLM-based embedders \citep{behnamghader2024llm2vec}, represent data
as points in $\mathbb{R}^d$ with similarity encoded by an inner product. The same approach dominates multimodal representation learning
\citep{radford2021learning, oquab2023dinov2}. Recent work has explored adapting the dimension of these representations \citep{kusupati2022matryoshka} or replacing them for binary codes \citep{gyurek2024binder}, but each concept still maps to a single point. In this section, we review the limitations of this paradigm and the alternatives that motivate our approach.

\paragraph{Limitations of vector representations.}
While powerful for capturing co-occurrence, vector spaces lack the intrinsic
structure to model asymmetric relations such as entailment or hierarchy
without external constraints \citep{park2024geometry, he2024language}.
Empirically, they are akin to bag-of-words representations
\citep{yuksekgonul2022and}, struggling to distinguish positive from negated
concepts \citep{gokhale2020vqa, singh2024learn, moreira2025learning,
alhamoud2025vision}. This ``negation blindness'' has spurred dedicated
benchmarks \citep{quantmeyer2024and, weller2023nevir,
zhang2025negvqavisionlanguagemodels}, on which standard dense retrievers
degrade significantly.

\paragraph{Hyperbolic embeddings.}
To capture hierarchy, hyperbolic embeddings \citep{nickel2017poincare,
nickel2018learning, ganea2018hyperbolic, bai2021modeling, xiong2022hyperbolic,
poppi2025hyperbolic} exploit the exponential growth of hyperbolic space to
embed trees compactly. While effective for pure taxonomies and transitive
inference, they impose a rigid ``continuous tree'' prior and may struggle with
non-hierarchical relations
\citep{sala2018representation, moreira2024hyperbolic}. They also lack native
operations for logical negation or conjunction, and require Riemannian
optimization and specialized indexing, limiting their scalability.

\paragraph{Region-based and probabilistic embeddings.}
A complementary line of work maps datapoints to regions rather than points:
partial order embeddings \citep{vendrov2015order}, Gaussians
\citep{vilnis2014word, athiwaratkun2018hierarchical}, beta distributions
\citep{ren2020beta}, box lattices \citep{vilnis2018probabilistic,
li2018smoothing, ren2020query2box}, and entailment cones
\citep{ganea2018hyperbolic_cones, dhall2020hierarchical, zhang2021cone}.
Boxes in particular support intersection and have proven effective for
modeling conjunctions, motivating our experimental comparison
in \S\ref{sec:experiments}. Crucially, however, none of these families are
closed under all of conjunction, disjunction, and negation, a property that
follows naturally from subspace geometry. These approaches also abandon the linear
structure of $\mathbb{R}^d$, forfeiting compatibility with standard
inner-product retrieval pipelines.

\paragraph{Quantum embeddings.}
Closest to our approach are quantum embeddings \citep{garg2019quantum,
srivastava2020inductive}, which represent concepts via subspaces, parameterized via orthogonal projection matrices. These methods enforce projection constraints through \emph{ad hoc} multi-step optimization, which complicates training and integration with
modern losses. Our differentiable relaxation removes this overhead, enabling
subspace learning with standard classification or contrastive objectives,
while remaining compatible with Euclidean retrieval.
\section{Subspace Representations}
\label{sec:method}
In this paper, rather than representing a datapoint as a single vector $\vx\in\sR^d$, we represent it as a region, instantiated as a subspace $\mathcal{S} \subseteq \sR^d$. As illustrated in Fig.~\ref{fig:overview}, instead of embedding the concept ``man on a boat'' as a single direction, we map it to the subspace $\mathcal{S}_\text{man on a boat}=\mathrm{span}(\{\vx_i\}_{i=1}^n)$. Each direction in $\mathcal{S}$ encodes a variation of the underlying concept: $\vx_1$ might represent a ``man on a boat that is fishing'' while $\vx_2$ a ``man on a boat that is not fishing''. The concept thus encompasses all instances that align with each $\vx_i$, or any linear combination thereof, representing the space of all possible instances \citep{van2004geometry, ganter2024formal}.

Formally, we parameterize a subspace $\mathcal{S}$ as the span of $n\geq d$ latent vectors $\mX=\begin{bmatrix}\vx_1 & \dots & \vx_n\end{bmatrix}\in\sR^{d\times n}$. If we let the thin singular value decomposition of $\mX$ be $\mU\mathbf{\Sigma}\mV^\top$, with $\mU\in\sR^{d\times r}$ and $\mU^\top\mU = \mI_r$, then $\mU$ is an orthonormal basis for the rank-$r$ subspace $\mathcal{S}$ spanned by the columns of $\mX$. We can write an equivalent representation of $\mathcal{S}$ through its orthogonal projection operator, 
\begin{equation}
    \mP := \mX(\mX^\top \mX)^{\dagger} \mX^\top = \mU\mU^\top \in \sR^{d \times d},
    \label{eq:hard_projector}
\end{equation}
where $\dagger$ is the pseudoinverse. This projector is symmetric ($\mP^\top = \mP$), idempotent ($\mP^2 = \mP$), and its trace equals the rank of $\mathcal{S}$,
\begin{equation}
    \mathrm{Tr}(\mP)= \mathrm{Tr}(\mU^\top\mU) = \mathrm{Tr}(\mI_r) = r.
    \label{eq:rank}
\end{equation} 

\paragraph{Subspace similarity and inclusion.} Cosine similarity between vectors can be generalized to subspaces $\mathcal{S}_i$, $\mathcal{S}_j$, with orthonormal bases $\mU_i$, $\mU_j$, respectively, via their projection operators $\mP_i$ and $\mP_j$, 
\begin{equation}
    \mathrm{sim}(\mP_i, \mP_j) := \mathrm{Tr}(\mP_i \mP_j) = \|\mU_i^\top\mU_j\|_F^2 = \sum_{k=1}^m \cos^2(\theta_k),
    \label{eq:similarity}
\end{equation}
where $\{\theta_k\}_{k=1}^m$ are the \emph{principal angles} between $\mathcal{S}_i$ and $\mathcal{S}_j$ and $m=\min\{\mathrm{rank}(\mathcal{S}_i),\mathrm{rank}(\mathcal{S}_j)\}$. Each $\theta_k$ is the smallest possible angle between a unit vector in $\mathcal{S}_i$ and a unit vector in $\mathcal{S}_j$, subject to orthogonality constraints on previously chosen directions. Thus, $\mathrm{sim}(\mP_i, \mP_j)$ measures the total squared alignment across the $m$ most comparable directions of the two subspaces, or their degree of \textit{overlap}. This recovers the squared cosine similarity as a special case: if $\mP_i$ and $\mP_j$ are rank-one projectors onto unit vectors $\vx_i$ and $\vx_j$, then $\mathrm{sim}(\mP_i,\mP_j) = (\vx_i^\top\vx_j)^2 = \cos^2(\angle(\vx_i,\vx_j))$.

An immediate consequence of Eqs. (\ref{eq:rank}) and (\ref{eq:similarity}) is that we can quantify subspace inclusion via a normalized inclusion score (NIS) based on the score from \citet{da2009normalized}:
\begin{equation}
    \mathrm{NIS}({\mP}_j \mid {\mP}_i) := \frac{\mathrm{sim}(\mP_i, \mP_j)}{\mathrm{Tr}(\mP_i)} \in [0,1].
    \label{eq:nis}
\end{equation}
This score attains 1 if and only if subspace $i$ is contained within subspace $j$. This formulation allows for an intuitive interpretation as a Bayes-like conditional probability: the probability of an instance belonging to subspace $j$ given it belongs to $i$.

\subsection{Algebraic Structure of Subspaces}
\label{sec:algebraic_structure_of_subspaces}
The power of subspaces lies in their structure, which natively supports interpretable operations between concepts. Using projection operators lets us map logical relations such as conjunction ($\land$), disjunction ($\lor$), and negation $(\neg)$ into the subspace operations of intersection $(\cap)$, linear sum $(+)$, and orthogonal complement $(\perp)$, respectively. These have linear-algebraic representations, addressing the limitations of vector embeddings discussed in \S\ref{sec:intro}.

\paragraph{Conjunction ($i\land j$).} Corresponds to the intersection of subspaces $\mathcal{S}_{i\land j} = \mathcal{S}_i \cap \mathcal{S}_j$. Any vector in $\mathcal{S}_{i\land j}$ is an element of $\mathcal{S}_i$ and $\mathcal{S}_j$. The product $\mP_i\mP_j$ is an orthogonal projection onto $\mathcal{S}_i\cap\mathcal{S}_j$ if and only if $\mP_i$ and $\mP_j$ commute. In the general case, $\mP_{i\land j} = \lim_{k \to \infty} (\mP_i \mP_j)^k$. Crucially, if concepts $i$ and $j$ are semantically disjoint, their subspace intersection collapses to the origin ($\{\mathbf{0}\}$). In Fig.~\ref{fig:overview}, the intersection of $\mathcal{S}_\text{man on a boat} = \mathrm{span}(\vx_1,\vx_2)$ and $\mathcal{S}_\text{is fishing} = \mathrm{span}(\vx_1,\vx_3)$ yields $\mathcal{S}_\text{man fishing on a boat} = \mathrm{span}(\vx_1)$.

\paragraph{Disjunction $(i \lor j)$.} Corresponds to the span (linear sum) of subspaces: $\mathcal{S}_{i\lor j} = \mathcal{S}_i + \mathcal{S}_j$. Any vector in $\mathcal{S}_{i\lor j}$ is a linear combination of elements in $\mathcal{S}_i$ or in $\mathcal{S}_j$. For commuting subspaces, the projection onto $\mathcal{S}_{i\lor j}$ satisfies $\mP_{i \lor j} = \mP_i + \mP_j - \mP_{i\land j}$. In Fig.~\ref{fig:overview}, the linear sum of $\mathcal{S}_\text{man fishing on a boat} = \mathrm{span}(\vx_1)$ and $\mathcal{S}_\text{man fishing not on a boat} = \mathrm{span}(\vx_3)$ yields $\mathcal{S}_\text{man fishing} = \mathrm{span}(\vx_1,\vx_3)$. 

\paragraph{Negation $(\neg i)$.} Corresponds to the subspace of all vectors orthogonal to the $\mathcal{S}_i$: $\mathcal{S}_{\neg i}=\mathcal{S}_i^\perp$. The projection operator onto $\mathcal{S}_i^\perp$ is given by $\mP_{\neg i} = \mI - \mP_i$. In Fig.~\ref{fig:overview}, the complement of $\mathcal{S}_\text{man on a boat}=\mathrm{span}(\vx_1,\vx_2)$ is given by $\mathcal{S}^\perp_\text{man on a boat}=\mathrm{span}(\vx_3)$.

\subsection{Subspaces as Soft Projection Operators}
\label{sec:soft_projection_operators}
While the orthogonal projector from Eq. (\ref{eq:hard_projector}) offers a rich and interpretable representation of a subspace, its optimization poses a challenge for gradient-based methods. Since the rank of a subspace is integer-valued, the space of all subspaces (a union of Grassmannian manifolds) is stratified and non-differentiable across rank changes. This makes it hard to simultaneously learn orientation and dimensionality via gradient-based methods. 

\paragraph{Soft projection operators.} To overcome the challenges associated with learning adaptive-rank subspaces we introduce a relaxation of the projection operator in Eq. (\ref{eq:hard_projector}). For a rank-$r$ subspace $\mathcal{S}$ spanned by the columns of $\mX \overset{\text{SVD}}{=}\mU\mathbf{\Sigma}\mV^\top\in\mathbb{R}^{d\times n}$, where $\mathbf{\Sigma}=\mathrm{diag}(\{\sigma_i\}_{i=1}^r)$ and $\mU\in\mathbb{R}^{d\times r}$ is the orthonormal basis of $\mathcal{S}$, we define a soft projector via Tikhonov regularization
\begin{equation}
    \tilde{\mP} := \mX(\mX^\top \mX + \lambda\mI)^{-1} \mX^\top = \mU \mathrm{diag}\left(\left\{\frac{\sigma_i^2}{\sigma_i^2 + \lambda}\right\}_{i=1}^{r}\right) \mU^\top,\quad\lambda>0.
    \label{eq:soft_projector}
\end{equation}
The hyperparameter $\lambda$ acts as a temperature: unlike a true projector ($\mP^2=\mP$), the eigenvalues of $\tilde{\mP}$ vary smoothly in $[0,1)$ rather than being binary. This makes the operator differentiable with respect to both orientation and rank, enabling gradual changes in dimensionality. Geometrically, this relaxation replaces the stratified manifold of projectors with a smooth manifold of positive semidefinite operators.

For small values of $\lambda$, the soft projectors in Eq.~(\ref{eq:soft_projector}) provide surrogates for the algebraic operations and metrics introduced in \S\ref{sec:algebraic_structure_of_subspaces}. The approximation error depends primarily on the weakest nonzero singular value $\sigma_r$ of $\mX$ and is upper-bounded by (see Appendix \ref{app:sec:theoretical_bounds})
\begin{equation}
    \epsilon(\sigma_r,\lambda) =\lambda/(\sigma_r^2 + \lambda).
\end{equation}
As $\lambda\!\to\!0$, $\epsilon(\sigma_r,\lambda)\!\to\!0$ and we recover the orthogonal projection operator $\tilde{\mP}\to\mP$, while larger $\lambda$ enforces smoother, more regularized projectors. Table~\ref{tab:soft_projectors} summarizes how each operation is approximated using $\tilde{\mP}$ and the resulting deviation from $\mP$ ($\lambda=0$).

\begin{table*}[t]
    \caption{Soft approximations of projection operations derived from $\mX_i$ and $\mX_j$. Errors are in operator norm, except rank (relative absolute error). $\sigma_r$, $\eta_r$ are the weakest nonzero singular values of $\mX_i$ and $\mX_j$. $\epsilon(\sigma_r,\lambda)=\lambda/(\sigma_r^2+\lambda)$.}
    \centering
    \begin{tabular}{llcccc}
        \toprule
         & \textbf{Projector} & \textbf{Negation} & \textbf{Intersection} & \textbf{Linear sum} & \textbf{Rank} \\
        \midrule
        Exact &
        $\mP=\mX(\mX^\top\mX)^\dagger\mX^\top$ &
        $\mI - \mP$ &
        $\mP_i \mP_j$ &
        $\mP_i + \mP_j - \mP_i\mP_j$ &
        $\mathrm{Tr}(\mP)$ \\
        Soft &
        $\tilde{\mP}=\mX(\mX^\top\mX + \lambda\mI)^{-1}\mX^\top$ &
        $\mI - \tilde{\mP}$ &
        $\tilde{\mP}_i \tilde{\mP}_j$ &
        $\tilde{\mP}_i + \tilde{\mP}_j - \tilde{\mP}_i\tilde{\mP}_j$ &
        $\mathrm{Tr}(\tilde{\mP})$ \\
        Error &
        $\epsilon(\sigma_r,\lambda)$ &
        $\epsilon(\sigma_r,\lambda)$ &
        $\epsilon(\sigma_r,\lambda)+\epsilon(\eta_r,\lambda)$ &
        $2(\epsilon(\sigma_r,\lambda)+\epsilon(\eta_r,\lambda))$ &
        $\epsilon(\sigma_r,\lambda)$ \\
        \bottomrule
    \end{tabular}
    \label{tab:soft_projectors}
\end{table*}

\begin{figure}
    \centering
    \includegraphics[width=0.9\linewidth]{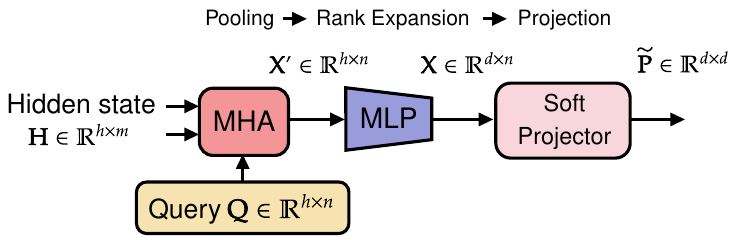}
    \caption{Subspace Projection Head (SPH).}
    \label{fig:sph}
\end{figure}

\paragraph{Subspace Projection Head (SPH)} To enable inductive subspace representations, we introduce the \emph{Subspace Projection Head (SPH)}, which converts any off-the-shelf transformer backbone into a subspace encoder. A transformer, augmented with a set of $n$ learnable query parameters $\mQ \in\mathbb{R}^{h\times n}$, first encodes text inputs into a contextualized hidden state $\mH\in\sR^{h \times m}$ (where $m$ is sequence length, $h$ is hidden dimension). The SPH transforms this hidden state $\mH$ into a fixed-size set of $n$ latent vectors $\mX\in\sR^{d\times n}$ that span a subspace $\mathcal{S}$ and then we compute the corresponding soft projector $\tilde{\mP}$. The SPH, depicted in Fig.~\ref{fig:sph}, comprises three stages:
\begin{enumerate}
    \item \textbf{MHA pooling:} The queries $\mQ$, finetuned with the rest of the transformer's parameters, attend to $\mH$ (acting as keys and values) via Multi-Head Attention (MHA), pooling $n$ vectors $\mX' = \mathrm{MHA}(\mathrm{query}=\mQ, \mathrm{key}=\mH, \mathrm{value}=\mH) \in\sR^{h\times n}$, akin to Set Transformer \citep{lee2019set}. This ensures the dimensions of $\mX'$ are independent of the sequence length.
    \item  \textbf{Rank expansion:} Since each head in the MHA outputs a linear combination of the columns of $\mH$, the rank of $\mX'$ is limited: $\mathrm{rank}(\mX') \leq m\cdot n_\text{heads}$. We address this via an MLP, which maps the $n$ vectors of dimension $h$ from the MHA output to $\sR^d$ as $\mX = \mathrm{MLP}(\mX')$. The non-linear activation in the MLP breaks the linear dependence from the attention mechanism, allowing the final matrix $\mX\in\sR^{d\times n}$ to achieve full rank $d$, regardless of the input sequence length.
    \item \textbf{Projection:} The soft projector representation $\tilde{\mP}$ onto $\mathcal{S}$ is computed from $\mX$ using the closed-form in Eq. (\ref{eq:soft_projector}).
\end{enumerate}

\subsection{Training Methodology}
We learn subspaces end-to-end via gradient descent, requiring no special pretraining or training constraints. Depending on the downstream task, we employ one of the following loss functions. 

\paragraph{Hierarchy reconstruction.} For similarity-based tasks \textit{e.g.}, \textsc{WordNet} reconstruction, we parameterize graph nodes with matrices $\mX_i$ and use an InfoNCE loss \citep{oord2019representationlearningcontrastivepredictive}, with the subspace similarity computed via $\mathrm{sim}(\tilde{\mP}_i,\tilde{\mP}_j)$, from Eq. (\ref{eq:similarity}). 

\paragraph{Link prediction.} In link prediction tasks, we optimize the normalized inclusion score $\mathrm{NIS}(\tilde{\mP}_i\mid \tilde{\mP}_j)$ from Eq. (\ref{eq:nis}) directly and use the margin loss from \citet{vendrov2015order}
\begin{equation}
    L = \sum_{(i,j)\in\mathcal{P}}[\gamma_+ - \mathrm{NIS}(\tilde{\mP}_i\mid \tilde{\mP}_j)]_+ + \sum_{(i,j)\in\mathcal{N}}[\mathrm{NIS}(\tilde{\mP}_i\mid \tilde{\mP}_j) - \gamma_-]_+,
    \label{eq:margin_loss}
\end{equation}
where $[\cdot]_+$ denotes the ReLU function. Here, $\gamma_+, \gamma_- \in (0,1)$ are the positive and negative margins and $\mathcal{P}$ and $\mathcal{N}$ the set of positive and negative nodes, respectively. 

\paragraph{NLI classification.} Textual entailment presents a unique challenge, requiring not just a measure of inclusion but also an explicit model of neutrality. For a  premise $p$ and hypothesis $h$, we model the relation $Y\in\{E,N,C\}$ (entailment, neutral, contradiction) as a discrete latent variable. For $Y\in\{E,C\}$, we assume the generative process for $S = \mathrm{NIS}(\tilde{\mP}_h\mid \tilde{\mP}_p)$
\begin{equation}
    S \mid (Y=y) \sim \mathrm{Beta}(\alpha_y, \beta_y), \quad y \in \{E, C\},
\end{equation}
with $\alpha_y \leq \beta_y$ if $y=C$ and $\beta_y \leq \alpha_y$ if $y=E$. We choose a Beta distribution because its support is $[0,1]$, matching the range of the NIS. For neutrals, subspace inclusion does not provide a reliable signal. Instead, we model neutrality by an MLP as
\begin{align}
    &P(Y = y \mid \tilde{\mP}_p, \tilde{\mP}_h) := \sigma\left(\mathrm{MLP}\left(\tilde{\mP}_p, \tilde{\mP}_h, \tilde{\mP}_p \tilde{\mP}_h, \tilde{\mP}_h \tilde{\mP}_p\right)\right),\quad y=N
    \label{eq:neutral_probability}
\end{align}
where $\sigma(\cdot)$ denotes the sigmoid function. Assuming uniform priors for entailment and contradiction classes, conditional on non-neutrality, we compute posterior probabilities for $y=E$ and $y=C$, denoted $P(Y=y \mid S=s, Y \in\{E,C\})$. The final posterior probabilities for $y\in\{E,C\}$ are then derived by combining the MLP output for neutrality with the Beta posteriors for non-neutrality:
\begin{align}
    &P(Y=y \mid \tilde{\mP}_p, \tilde{\mP}_h,S=s) \nonumber \\ &= (1-P(Y=N \mid \tilde{\mP}_p, \tilde{\mP}_h))
    P(Y=y \mid S=s, Y \neq N), 
    \label{eq:non_neutral_probability}
\end{align}
for  $y \in \{E,C\}$. The posteriors in Eqs. (\ref{eq:neutral_probability}) and (\ref{eq:non_neutral_probability}) are optimized via a cross-entropy loss.

A key insight into how these losses shape the subspaces is revealed by the gradient of the subspace similarity (\ref{eq:similarity}). As derived in Appendix~\ref{app:sec:training_dynamics}, $\nabla_{\mX_i}\mathrm{sim}(\tilde{\mP}_i,\tilde{\mP}_j)$ encourages subspace $i$ to expand along the principal directions of subspace $j$ that it lacks. This promotes subspace inclusion, and the gradient vanishes once one subspace is contained within the other, leading to stable convergence.

\paragraph{Efficiency considerations.} While computing $\tilde{\mP}$ from $\mX\in\mathbb{R}^{d\times n}$ is $\mathcal{O}(n^3)$ in the number of vectors $n$, and has a memory footprint that scales with $d^2$, where $d$ is the ambient dimension, our approach is practical for two key reasons. First, we learn a data-dependent rank for each subspace. As our experiments demonstrate, this allows for considerable compression of $\tilde{\mP}$ via low-rank approximations. Second, the subspace similarity (\ref{eq:similarity}) and NIS (\ref{eq:nis}) are equivalent to dot products between the vectorized matrices: $\mathrm{Tr}(\tilde{\mP}_i \tilde{\mP}_j) = \mathrm{vec}(\tilde{\mP}_i)^\top\mathrm{vec}(\tilde{\mP}_j)$. This allows our subspaces to be indexed by highly optimized vector search libraries, making large-scale retrieval feasible.

\section{Experiments}
\label{sec:experiments}
We empirically validate our subspace embeddings on three axes: hierarchy modeling on \textsc{WordNet} \citep{miller1995wordnet} (reconstruction in \S\ref{sec:wordnet_reconstruction} and link prediction in \S\ref{sec:wordnet_linkprediction}), entailment on SNLI \citep{bowman2015large} in \S\ref{sec:snli} and compositional entailment in \S\ref{sec:compositional_entailment}.

\subsection{WordNet Reconstruction}
\label{sec:wordnet_reconstruction}
We first assess the capacity of our representations to encode known hierarchical relations. In \textsc{WordNet}'s reconstruction task, all edges from the full transitive closure of the noun and verb hypernymy hierarchies are used for training and testing.
 
\paragraph{Experimental details.} Each node in the graph is represented by a soft projection matrix $\tilde{\mP}_i$ from Eq. (\ref{eq:soft_projector}), with $\lambda=0.2$, parameterized by a matrix $\mX_i\in\mathbb{R}^{d\times d}$. We considered two ambient space dimensions, $d=64$ and $d=128$, setting the number of vectors as $n=d$ in each case. For each training edge $(u,v)$, we sample 19 nodes $v' \neq u$ such that neither $(u,v')$ nor $(v',u)$ are in the train split and optimize InfoNCE using Adam \citep{kingma2017adammethodstochasticoptimization}. During evaluation, we first compute the subspace similarity $\mathrm{Tr}(\tilde{\mP}_u\tilde{\mP}_v)$ of every edge $(u,v)$ in the transitive closure. We then rank each of these scores among those of all node pairs that are not connected in the transitive closure. Based on these rankings, we report the mean rank (MR) and the mean average precision (mAP), following \citet{nickel2017poincare}. We compare against Euclidean vectors in $\mathbb{R}^{128}$ and hyperbolic embeddings with $d=10$, as these representations are known to achieve optimal efficiency in low dimensions. Additional details in Appendix \ref{app:sec:wordnet_reconstruction}.

\begin{table}[t]
    \caption{\textsc{WordNet} reconstruction results. mAP = Mean Average Precision, MR = Mean Rank, $\rho$ = Spearman's rank correlation with taxonomy rank.}
    \centering
    \small
    \begin{tabular}{lcccccccc}
        \toprule
        \multirow{2}{*}{} & \multicolumn{4}{c}{\textbf{Nouns}} & \multicolumn{4}{c}{\textbf{Verbs}} \\
        \cmidrule(lr){2-5} \cmidrule(lr){6-9}
        & mAP ($\uparrow$) & MR ($\downarrow$) & $\rho$ ($\uparrow$) &  &
          mAP ($\uparrow$) & MR ($\downarrow$) & $\rho$ ($\uparrow$) &  \\
        \midrule
        $\mathbb{R}^{128}$ 
            & 95.1 & 1.31 & -- & & 98.6 & 1.04 & -- & \\
        $\mathcal{P}^{10}$
            & 86.5 & 4.02 & 58.5 & & 91.2 & 1.35 & 55.1 & \\
        $\mathcal{H}^{10}$
            & 92.8 & 2.95 & 59.5 & & 93.3 & 1.23 & 56.6 & \\
        \cellcolor{highlightblue!40}SE$^{64}$
            & \cellcolor{highlightblue!40}97.8 
            & \cellcolor{highlightblue!40}1.06 
            & \cellcolor{highlightblue!40}65.9 
            & \cellcolor{highlightblue!40} & 
            \cellcolor{highlightblue!40}99.5 
            & \cellcolor{highlightblue!40}1.01  
            & \cellcolor{highlightblue!40}65.8 & \cellcolor{highlightblue!40} \\
         \cellcolor{highlightblue!40}SE$^{128}$
            & \cellcolor{highlightblue!40}\textbf{98.6} 
            & \cellcolor{highlightblue!40}\textbf{1.04} 
            & \cellcolor{highlightblue!40}\textbf{68.0} 
            & \cellcolor{highlightblue!40} & 
            \cellcolor{highlightblue!40}\textbf{99.9} 
            & \cellcolor{highlightblue!40}\textbf{1.00}  
            & \cellcolor{highlightblue!40}\textbf{67.0} & \cellcolor{highlightblue!40} \\
        \bottomrule
    \end{tabular}
    \label{tab:wordnet_reconstruction}
\end{table}

\paragraph{Reconstruction results.} Our method achieves state-of-the-art performance on \textsc{WordNet} reconstruction. As shown in Table \ref{tab:wordnet_reconstruction}, our subspace representations (SE$^{64}$ and SE$^{128}$) significantly outperform both Hyperbolic (Poincaré $\mathcal{P}^{10}$ and Lorentz $\mathcal{H}^{10}$ models) and Euclidean ($\mathbb{R}^{128}$) baselines, achieving a near-perfect reconstruction on the shallower verb hierarchy. Further, the learned effective rank, computed as $\operatorname{Tr}(\tilde{\mathbf{P}}_i)$, shows a strong Spearman's rank correlation ($\rho$) with the taxonomic rank, suggesting subspace dimension acts as an emergent measure of specificity.

\paragraph{Generalization to graded lexical entailment.} To assess how our \textsc{WordNet} representations generalize to graded lexical entailment, we evaluated them zero-shot on the \textsc{HyperLex} noun subset, which contains human-annotated entailment scores (0--10) between 2,163 noun pairs (see Appendix \ref{app:sec:hyperlex}). We quantify entailment using the NIS from Eq. (\ref{eq:nis}), selecting the \textsc{WordNet} synset pair with maximal similarity for disambiguation \citep{athiwaratkun2018hierarchical}. As reported in Table \ref{tab:hyperlex}, our embeddings show a significantly stronger correlation with human judgments than prior work, achieving a Spearman's rank correlation of 0.683, substantially outperforming Euclidean, Poincaré, and Gaussian embedding baselines.

\begin{table}[t]
    \caption{\textsc{HyperLex} Spearman's rank correlation with human-annotated entailment scores (\textsc{WordNet} embeddings).}
    \centering
    \small
    \begin{tabular}{lccc>{\columncolor{highlightblue!40}}c}
        \toprule
         & $\mathbb{R}^5$  &   $\mathcal{P}^5$ &  DOE-A$^{50}$ & SE$^{128}$ \\
        \midrule
        Spearman's correlation $(\uparrow)$ & 0.389 & 0.512 & 0.590 & \textbf{0.683}  \\
        \bottomrule
    \end{tabular}
    \label{tab:hyperlex}
\end{table}

\vspace{-2pt}
\subsection{WordNet Link Prediction}
\label{sec:wordnet_linkprediction}
In the link prediction task, we assess generalization from sparse supervision. We split the set of transitive closure edges that are not part of the original graph (non-basic edges) into train (90\%), validation (5\%) and test (5\%) using the split from \citet{suzuki2019hyperbolic}. 

\paragraph{Experimental details.} To assess how the percentage of the transitive closure seen during training impacts performance, we created partial training edge coverages by randomly sampling 0\%, 10\%, 25\% or 50\% of non-basic edges, to which we append all the basic edges. We considered two ambient space dimensions, $d=64$ and $d=128$, setting the number of vectors as $n=d$ in each case. Training was performed by optimizing the margin loss defined in Eq. (\ref{eq:margin_loss}). During evaluation, for each positive test edge, we consider 10 negative test edges: half with a corrupted head, and half with a corrupted tail. We classify edges by thresholding the NIS from Eq. (\ref{eq:nis}) and report the classification F1 score averaged over 5 random seeds.

\paragraph{Link prediction results.} Link prediction results are shown in Table \ref{tab:wordnet_linkprediction}. We compare against Euclidean embeddings ($\sR^{10}$), order embeddings (OE$^{10}$) \cite{vendrov2015order}, Poincaré ($\mathcal{P}^{10}$) \cite{nickel2017poincare}, hyperbolic entailment cones (Cones$^{10}$) \citep{ganea2018hyperbolic_cones}, hyperbolic disk embeddings (Disk$^{10}$) \citep{suzuki2019hyperbolic} and umbral half-space embeddings (UHS$^{10}$) \citep{yushadow}. Subspaces are most effective when supervision is
sparse: SE$^{128}$ leads at 0\%, 10\%, and 25\% coverage, improving over the
next-best method by 4.9 F1 points at 10\%. At 50\% coverage, all leading
methods saturate above 95 F1, with UHS slightly ahead.

\begin{table}[t]
    \caption{\textsc{WordNet} noun \textbf{link prediction} F1 scores ($\uparrow$). Superscript denotes dimension.}
    \centering
    \begin{tabular}{lcccc}
    \toprule
    Model & 0\% & 10\% & 25\% & 50\% \\
    \midrule
    $\sR^{10}$        & 29.4 & 75.4 & 78.4 & 78.1 \\
    OE$^{10}$         & 43.0 & 69.7 & 79.4 & 84.1 \\
    $\mathcal{P}^{10}$ & 29.0 & 71.5 & 82.1 & 85.4 \\
    Cones$^{10}$      & 32.4 & 84.9 & 90.8 & 93.8 \\
    Disk$^{10}$       & 36.5 & 79.5 & 90.5 & 94.2 \\
    UHS$^{10}$        & 52.2 & 89.4 & 95.7 & \textbf{97.0} \\
    \cellcolor{highlightblue!40}SE$^{64}$  
                       & \cellcolor{highlightblue!40}$49.0\pm0.11$
                       & \cellcolor{highlightblue!40}$93.6\pm0.06$
                       & \cellcolor{highlightblue!40}$95.9\pm0.08$
                       & \cellcolor{highlightblue!40}$95.8\pm0.07$ \\
    \cellcolor{highlightblue!40}SE$^{128}$ 
                       & \cellcolor{highlightblue!40}$\mathbf{53.4\pm0.41}$
                       & \cellcolor{highlightblue!40}$\mathbf{94.3\pm0.05}$
                       & \cellcolor{highlightblue!40}$\mathbf{95.9\pm0.11}$
                       & \cellcolor{highlightblue!40}$95.5\pm0.06$ \\
    \bottomrule
    \end{tabular}
    \label{tab:wordnet_linkprediction}
\end{table}

\subsection{SNLI}
\label{sec:snli}
We conducted experiments on NLI using the SNLI dataset. SNLI comprises 550,152 training and 10,000 validation/test premise ($p$) -- hypothesis ($h$) pairs, each annotated with one of three labels: entailment, neutral, or contradiction. We evaluate in 3-way and 2-way
(entailment vs.\ non-entailment) regimes, comparing three bi-encoder
approaches across two backbones (all-MiniLM-L6-v2, mpnet-base-v2). The first
is an MLP classifier in two variants: $\mathrm{MLP}(\vp, \vh)$ uses
concatenated embeddings; $\mathrm{MLP}(\vp, \vh, \vp-\vh)$ adds their
difference. The second is box embeddings \citep{li2018smoothing}, which map
each sentence to an axis-aligned box in $\mathbb{R}^d$ at the same ambient
dimension as SPH. Our model maps $p$ and $h$ to soft projectors via the SPH
(\S\ref{sec:soft_projection_operators}). Additional details are provided in Appendix~\ref{app:sec:nli_experiments}.

\begin{table}[t]
    \centering
    \setlength{\tabcolsep}{12pt}
    \caption{SNLI \textbf{test accuracy}: 2-way (entailment vs. non-entailment) and 3-way (entailment, neutral, contradiction).}
    \begin{tabular}{lcc}
        \toprule
        \textbf{Method} & \textbf{2-way} & \textbf{3-way} \\
        \midrule
        OE (GRU) & 88.60 & -- \\
        HNN (GRU) & 81.19 & -- \\
        \midrule
        \multicolumn{3}{l}{\textbf{all-miniLM-L6-v2 (22.7M params)}} \\
        MLP$(\vp,\vh)$ & 90.16 $\pm$ 0.19 & 83.57 $\pm$ 0.11 \\
        MLP$(\vp,\vh,\vp-\vh)$ & 91.02 $\pm$ 0.10 & 84.89 $\pm$ 0.21\\
        \rowcolor{highlightblue!40} SPH (SE$^{64}$) & 91.02 $\pm$ 0.11 & 84.43 $\pm$ 0.07  \\
        \rowcolor{highlightblue!40} SPH (SE$^{128}$) & \textbf{91.25} $\pm$ \textbf{0.09} & \textbf{85.41}$\pm$ \textbf{0.09} \\
        \midrule
        \multicolumn{3}{l}{\textbf{mpnet-base-v2 (109M params)}} \\
        MLP$(\vp,\vh)$ & 90.67 $\pm$ 0.33 & 84.10 $\pm$ 0.27 \\
        MLP$(\vp,\vh,\vp-\vh)$ & 91.74 $\pm$ 0.07 & 85.68 $\pm$ 0.15 \\ %
        Box Embeddings$^{64}$  & 91.25 $\pm$ 0.18 &  -- \\
        Box Embeddings$^{128}$ & 91.70 $\pm$ 0.09 &  -- \\
        \rowcolor{highlightblue!40} SPH (SE$^{64}$) & 91.77 $\pm$ 0.42 &  85.63 $\pm$ 0.14 \\
        \rowcolor{highlightblue!40} SPH (SE$^{128}$) &\textbf{91.91} $\pm$ \textbf{0.08} &  \textbf{85.80} $\pm$ \textbf{0.05} \\
        \bottomrule
    \end{tabular}
    \label{tab:snli_results}
\end{table}

\begin{figure}
    \centering
    \includegraphics[width=0.95\linewidth]{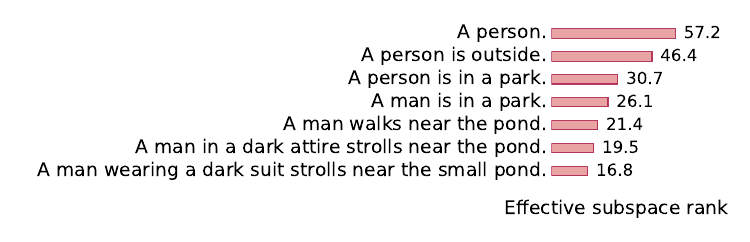}
    \includegraphics[width=0.95\linewidth]{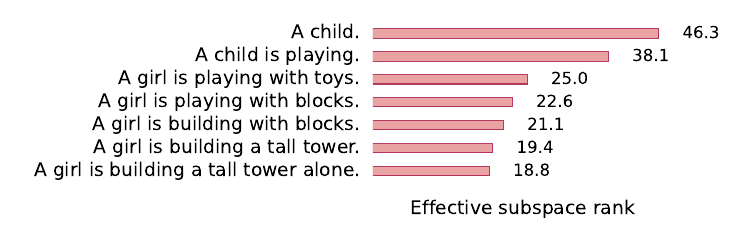}
    \includegraphics[width=0.95\linewidth]{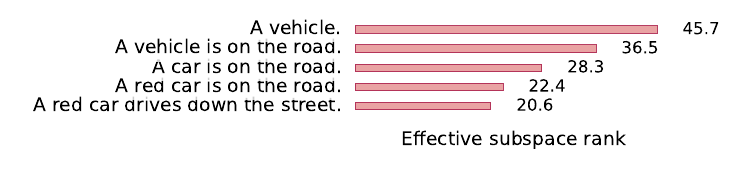}
    \caption{Effective subspace ranks for three entailment chains, encoded by mpnet-base-v2 + SPH (SE$^{128}$), increasing monotonically from specific descriptions to general ones.}
    \label{fig:entailment_chain}
\end{figure}

\paragraph{Results.} SNLI results are shown in
Table~\ref{tab:snli_results}. SE matches the strongest bi-encoder
baselines on standard 2-way and 3-way classification, slightly exceeding them
on 3-way. With this experiment, our goal is not to claim state-of-the-art on SNLI, but to verify that subspace embeddings provide a competitive geometric solution to a standard entailment task. The evaluation of whether subspaces provide additional capacity for compositionality is the subject of \S\ref{sec:compositional_entailment}.

\subsection{Compositional Entailment}
\label{sec:compositional_entailment}
To probe how the SNLI-trained embeddings from \S\ref{sec:snli} perform zero-shot under composite hypotheses, we generated 600 premise--composite-hypothesis
pairs (dataset details are provided in Appendix~\ref{app:sec:composite_entailment}). Given 150 premises,
each associated with two entailed atomic hypotheses ($h_1, h_2$) and one
contradicted hypothesis ($h_3$), we form four composites per premise: two
entailed ($h_1 \land h_2$, $h_1 \land \neg h_3$) and two contradicted
($h_1 \land h_3$, $h_1 \land \neg h_2$). For example, given $p = $
``Two children are sitting on a red picnic blanket, eating sandwiches'',
the composite ``People are eating'' $\land$ ``People sitting directly on
the grass'' is contradicted. As vector baselines lack native logical operators, we follow
common practice in using vector averaging for conjunction and vector difference for
negation \citep{mikolov2013distributed}. Subspaces and boxes both compose
via intersection; subspaces additionally use the orthogonal complement for
negation. We report binary-entailment AUC, scored by the NIS for subspaces,
its volumetric analog (intersection volume over premise volume) for boxes,
and the trained MLP probability for vector baselines.

\paragraph{Results.} Table~\ref{tab:compositional_entailment} shows that vector
baselines perform adequately on conjunction (83--91\% AUC) but fail
catastrophically on negation, dropping to near-random performance
(49--69\% AUC). In contrast, all subspace embeddings exceed 90\% AUC on
both operations, matching their atomic-hypothesis performance. Box
embeddings, which support intersection, achieve competitive
conjunction AUC (90--92\%) but lag our subspaces by 4--7 points
despite being purpose-built for this operation. More fundamentally, the
complement of a box is not itself a box, leaving no
closed-form negated representation within the model class. Our subspaces close this
expressivity gap, achieving $>$90\% AUC on negation while retaining
inner-product retrieval compatibility.

\begin{table}[]
    \caption{Zero-shot compositional entailment \textbf{ROC AUC}.}
    \centering
    \begin{tabular}{lcc}
        \toprule
         & \multicolumn{2}{c}{\textbf{AUC}} \\
        \textbf{Model} & Conj. ($\land$) & Neg. ($\land\neg$) \\
        \midrule
        all-MiniLM-L6-v2 + MLP($\vp,\vh$)                              & 86.45 & 57.68 \\
        all-MiniLM-L6-v2 + MLP($\vp,\vh$,$\vp-\vh$)                    & 91.22 & 48.62 \\
        mpnet-base-v2 + MLP($\vp,\vh$)                                 & 82.91 & 55.75 \\ 
        mpnet-base-v2 + MLP($\vp,\vh,\vp-\vh$)                         & 90.20 & 68.69 \\ 
        mpnet-base-v2 + Box$^{64}$                                     & 91.51 & -- \\ 
        mpnet-base-v2 + Box$^{128}$                                    & 89.95 & -- \\ 
        \rowcolor{highlightblue!40}all-MiniLM-L6-v2 + SPH (SE$^{64}$)  & 94.68 & 90.49 \\
        \rowcolor{highlightblue!40}all-MiniLM-L6-v2 + SPH (SE$^{128}$) & 95.02 & 92.77 \\
        \rowcolor{highlightblue!40} mpnet-base-v2 + SPH (SE$^{64}$)    & 95.87 & 93.89 \\
        \rowcolor{highlightblue!40} mpnet-base-v2 + SPH (SE$^{128}$)   & \textbf{96.55} & \textbf{95.76} \\
        \bottomrule
    \end{tabular}
    \label{tab:compositional_entailment}
\end{table}

\subsection{Efficiency Analysis}
\label{sec:efficiency_analysis}
\paragraph{Retrieval latency.} Given the embeddings for the 155,070 captions from the Flickr30k dataset \citep{young2014image}, we benchmarked top-10 retrieval latency on CPU using batches of 128 queries. We compared our subspace embeddings (SE$^{128}$) against a 10-dimensional Poincaré hyperbolic baseline ($\mathcal{P}^{10}$). Because hyperbolic distance is non-Euclidean, we applied brute-force search over the entire database, ranking by the negative hyperbolic distance. In contrast, our NIS score can be formulated as a maximum inner product search problem between query and caption vectors:
\begin{equation}
    \mathrm{NIS}(\tilde{\mP}_\text{query}\mid \tilde{\mP}_\text{caption}) = \left(\frac{\mathrm{vec}(\tilde{\mP}_\text{caption})}{\mathrm{Tr}(\tilde{\mP}_\text{caption})}\right) ^\top\mathrm{vec}(\tilde{\mP}_\text{query}).
\end{equation}
This formulation allows us therefore to use fast approximate search libraries. We indexed the normalized caption vectors using a CPU index from the FAISS library \cite{douze2024faiss}. We used an inverted file index with Product Quantization (IndexIVFPQ). The index was trained on 50,000 vectors. We used 64 subquantizers for PQ with 8 bits per subquantizer, and set the search-time parameter to $n_\text{probe}=32$. The results in Table~\ref{tab:retrieval_latency} show that, with FAISS, SE$^{128}$ is nearly 8$\times$ faster than a 10D Poincaré ($\mathcal{P}^{10}$) baseline.

\paragraph{Encoding time.} We also measured the overhead introduced by our SPH module when encoding Flickr30k captions on an RTX8000 GPU with 49GB of memory. To isolate the computational cost of the forward pass, tokenization (max-length of 35) and data transfers were computed beforehand. Table~\ref{tab:encoding_time} shows the average encoding time per query (forward-pass) for different batch sizes, demonstrating that the additional computational cost is modest, averaging an additional 0.12ms/query for a batch size of 128.

\begin{table}[]
    \caption{\textbf{GPU average encoding time (ms/query)}, averaged over Flickr30k's captions.}
    \centering
    \begin{tabular}{l c c c c c c}
        \toprule
         & \multicolumn{5}{c}{\textbf{Batch size}} \\
        \cmidrule(lr){2-6}
        \textbf{Model} & 1 & 4 & 16 & 64 & 128 \\
        \midrule
        mpnet-base-v2 & 5.95 & 1.74 & 0.69 & 0.59 & 0.56 \\ 
        \rowcolor{highlightblue!40} mpnet-base-v2 + SPH (SE$^{128}$) & 6.80 & 2.13 & 0.86 & 0.73 & 0.68 \\
        \bottomrule
    \end{tabular}
    \label{tab:encoding_time}
\end{table}

\begin{table}[]
    \centering
    \caption{Top-10 retrieval latency on Flickr30k captions.}
    \label{tab:retrieval_latency}
    \begin{tabular}{lc} 
        \toprule
         & \textbf{Latency (ms/query)} \\
        \midrule
        $\mathcal{P}^{10}$ (brute-force) & 3.6 \\
        \rowcolor{highlightblue!40} SE$^{64}$ (brute-force) & 0.9 \\
        \rowcolor{highlightblue!40} SE$^{128}$ (brute-force) & 3.4 \\
        \rowcolor{highlightblue!40} SE$^{128}$ (FAISS) & \textbf{0.5} \\
        \bottomrule
    \end{tabular}
\end{table}

\subsection{Qualitative Analysis}
\begin{figure}
    \centering
    \includegraphics[width=0.49\linewidth]{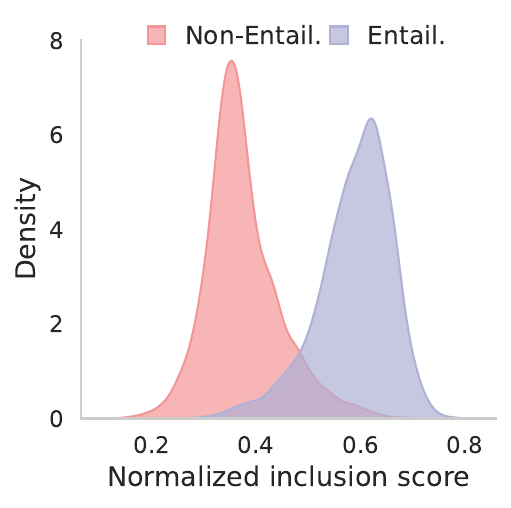}
    \includegraphics[width=0.49\linewidth]{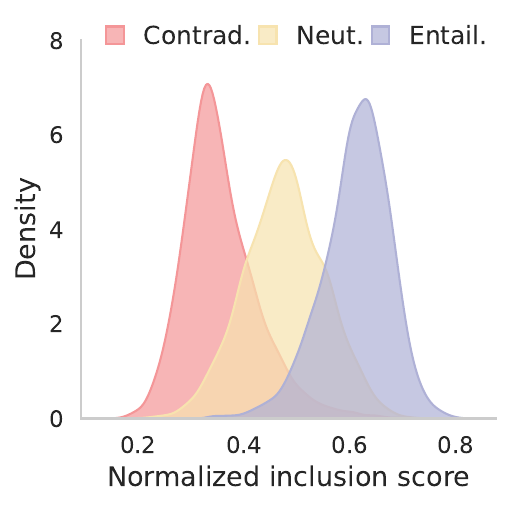}
    \captionof{figure}{Entailment as subspace inclusion: NIS histogram of SNLI's test set encoded with SE$^{128}$. Left: 2-way; Right: 3-way.} 
    \label{fig:histogram_nis_3way}
\end{figure}

A key finding of our work is that our framework learns an interpretable geometry that maps the hierarchical structure of language onto the representations. We confirm this empirically on SNLI, using our SE$^{128}$ embeddings. In Fig.~\ref{fig:histogram_nis_3way}, we plot the histogram of the NIS (\ref{eq:nis}) for premise-hypothesis pairs encoded with our mpnet-base-v2 (SE$^{128}$) subspace model. We observe that for entailment this metric is concentrated towards 1, for contradictions it skews towards 0, and for neutrals it is centered around 0.5. This confirms that the NIS reflects the underlying entailment structure via subspace inclusion: each premise subspace is approximately contained within the subspaces of the hypotheses it entails.

\begin{figure}
    \centering
    \includegraphics[width=0.9\linewidth]{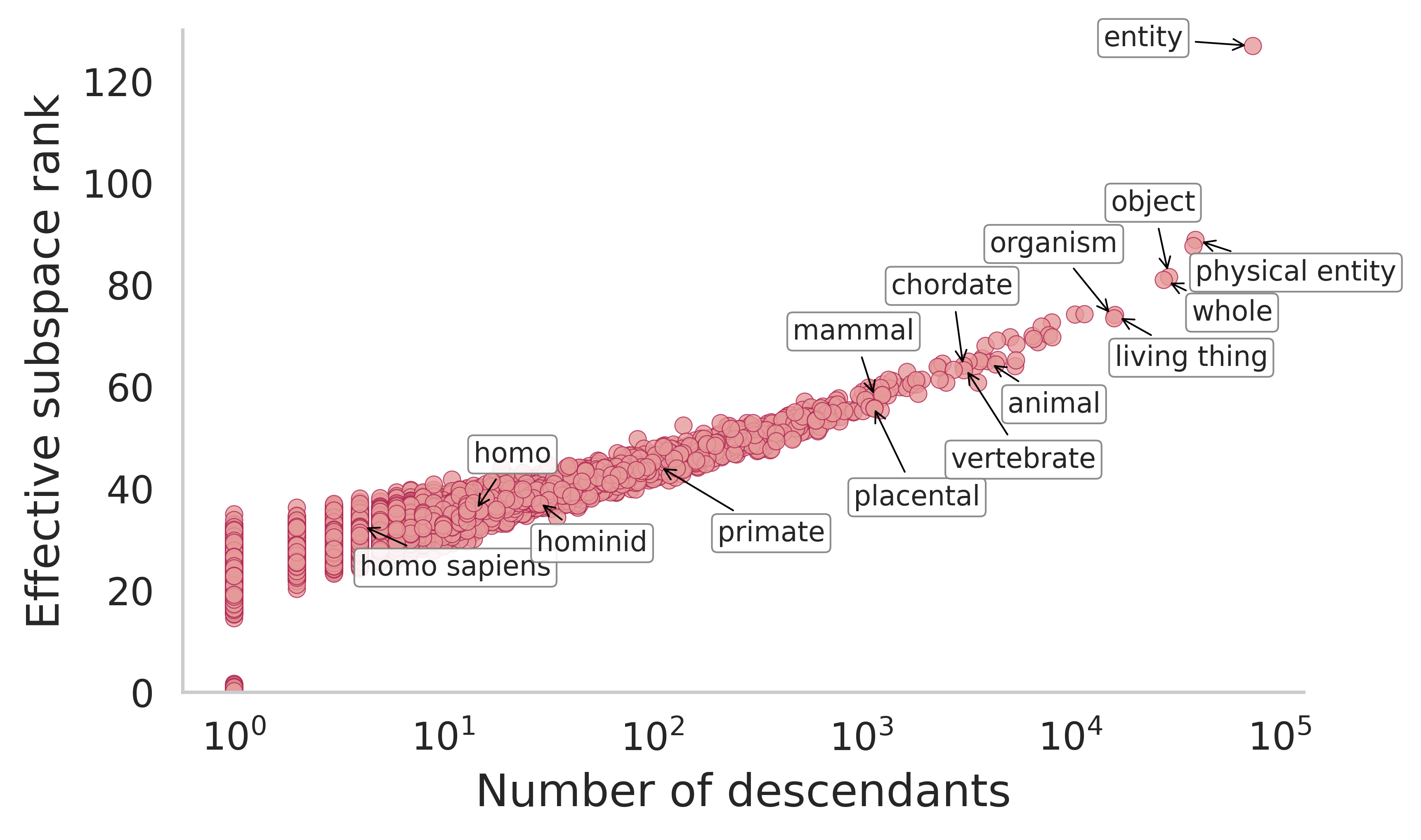}
    \caption{\emph{Effective rank} $\mathrm{Tr}(\tilde{\mP})$ of our $\mathrm{SE}^{128}$ embeddings vs. number of descendants of all \textsc{WordNet} nouns.}
    \label{fig:dim_x_deg}    
\end{figure}

\paragraph{Rank as a measure of generality.} A direct consequence of this is that a subspace's \emph{effective rank}, as measured by $\mathrm{Tr}(\tilde{\mP})$, becomes an emergent measure of semantic generality. For a specific concept to be nested within many broader ones, it must occupy a lower-dimensional subspace. This property is confirmed quantitatively by the high Spearman correlation ($\rho$) between \textsc{WordNet} nouns' true hierarchical positions (distance from root) and their learned \emph{effective rank} in Table~\ref{tab:wordnet_reconstruction}. We provide additional visual confirmation of this principle. Fig.~\ref{fig:dim_x_deg} shows how the \emph{effective rank} of \textsc{WordNet}'s nouns grows with the number of their descendants. The annotated chain from the specific \textit{homo sapiens} to the root noun \textit{entity} illustrates this monotonic increase. Fig.~\ref{fig:entailment_chain} shows the same phenomenon for three entailment sequences encoded with our SE$^{128}$ model. We observe again that the \emph{effective rank} of each sentence increases as we go from a specific description to a general one \textit{e.g.}, ``A red car drives down the street.'' $\rightarrow$ ``A vehicle is on the road.'' $\rightarrow$ ``A vehicle.''.

\begin{figure}
    \centering
    \includegraphics[width=0.9\linewidth]{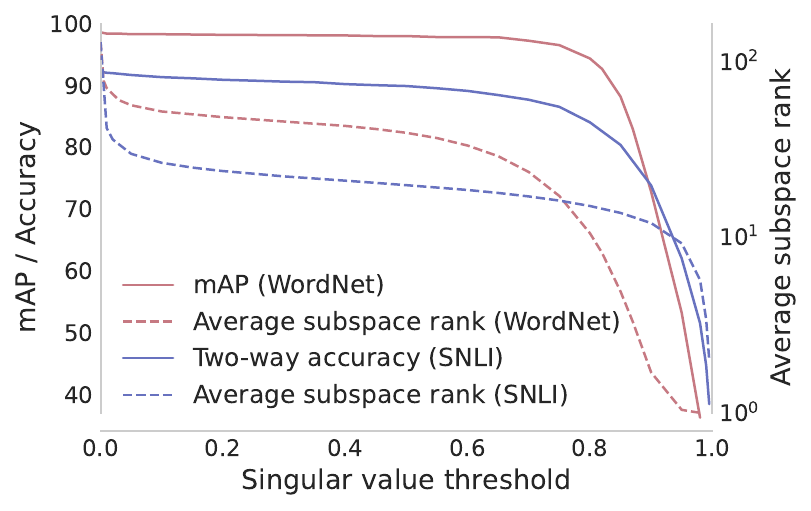}
    \caption{Accuracy, mAP and average rank as a function of the singular value threshold.}
    \label{fig:compression}
\end{figure}

\paragraph{Dimensionality reduction.} This learned structure, where the rank encodes specificity, makes our embeddings inherently compressible. Since each subspace dynamically allocates the dimensions needed to represent each concept, we can perform post-training compression via truncated SVD, with minimal performance loss. To assess this capability, we approximated \textsc{WordNet} and SNLI embeddings $\tilde{\mP}_i$ by retaining singular values greater than a threshold $\tau\in [0,1]$ and plotted the reconstruction mAP, in the case of \textsc{WordNet}, or the two-way accuracy, for SNLI, as well as the average subspace rank, as a function of $\tau$. As shown in Fig.~\ref{fig:compression}, the learned subspaces exhibit rapid spectral decay in both experiments, allowing for compression of up to $4\times$ with negligible impact on task performance. This paves the way for a new class of embeddings where representational complexity is not fixed, but a learned, data-driven property.


\begin{figure}
    \centering
    \begin{subfigure}{\linewidth}
        \includegraphics[width=\linewidth]{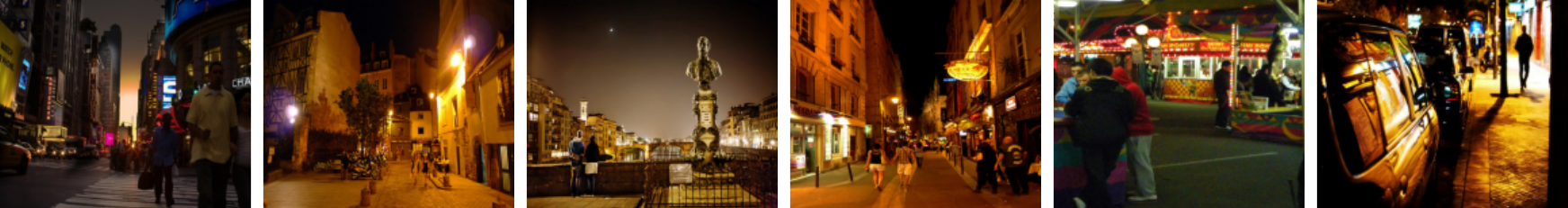}
        \caption{$\tilde{\mP}_{\text{a city street}}\tilde{\mP}_{\text{at night}}$}
    \end{subfigure}

    \begin{subfigure}{\linewidth}
        \includegraphics[width=\linewidth]{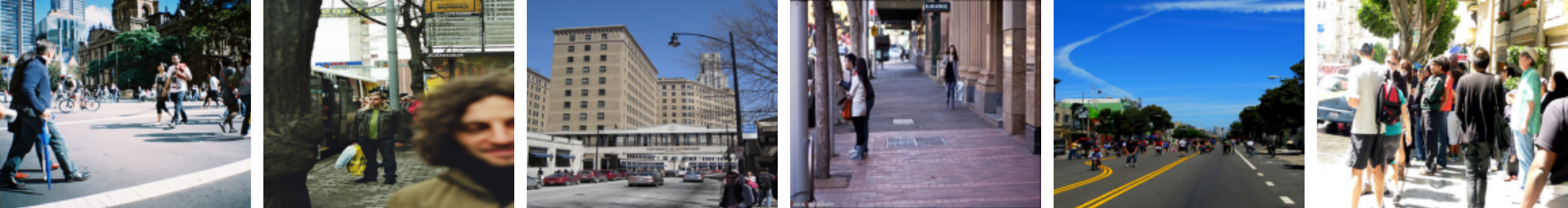}
        \caption{$\tilde{\mP}_{\text{a city street}}(\mI-\tilde{\mP}_{\text{at night}})$}
    \end{subfigure}

    \begin{subfigure}{\linewidth}
        \includegraphics[width=\linewidth]{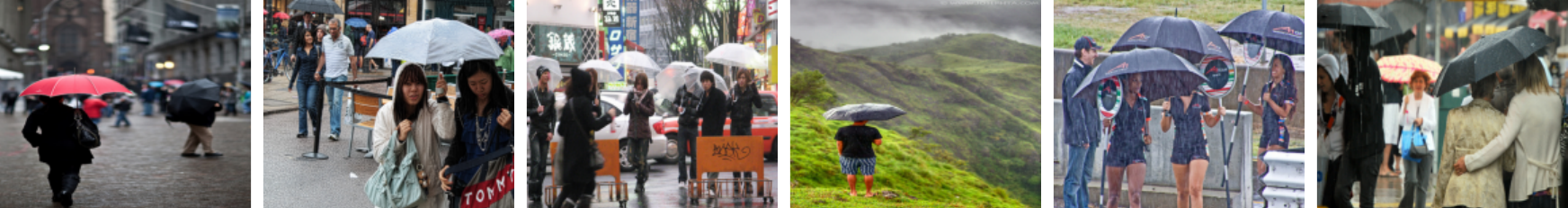}
        \caption{$\tilde{\mP}_{\text{an umbrella}}\tilde{\mP}_{\text{it’s raining}}$}
    \end{subfigure}

    \begin{subfigure}{\linewidth}
        \includegraphics[width=\linewidth]{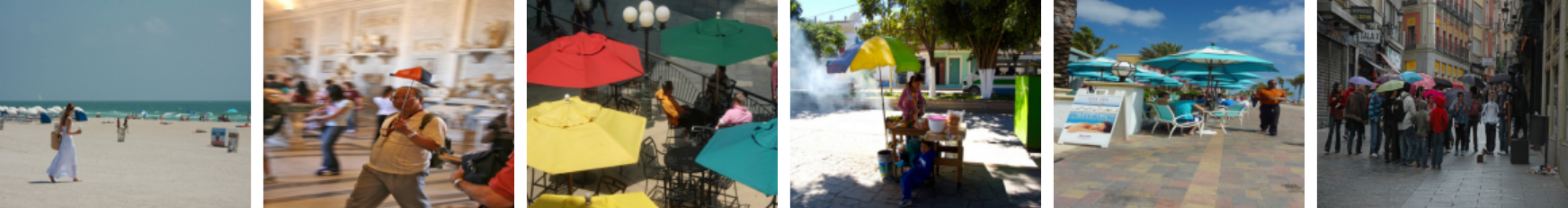}
        \caption{$\tilde{\mP}_{\text{an umbrella}}(\mI-\tilde{\mP}_{\text{it’s raining}})$}
    \end{subfigure}

    \begin{subfigure}{\linewidth}
        \includegraphics[width=\linewidth]{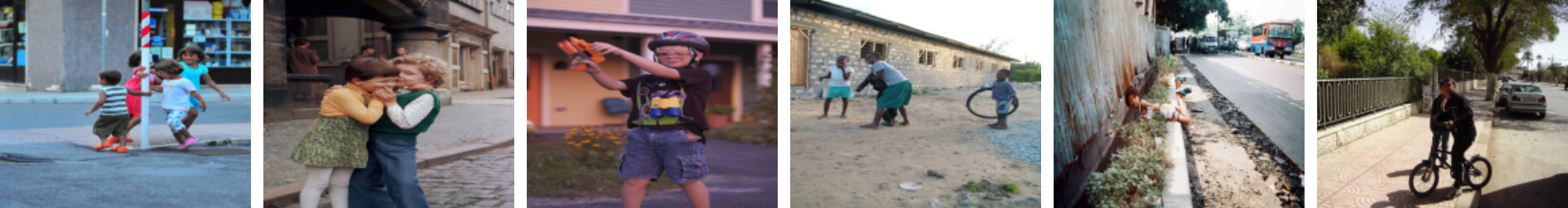}
        \caption{$\tilde{\mP}_{\text{a child playing outdoors}}\tilde{\mP}_{\text{near a road}}$}
    \end{subfigure}

    \begin{subfigure}{\linewidth}
        \includegraphics[width=\linewidth]{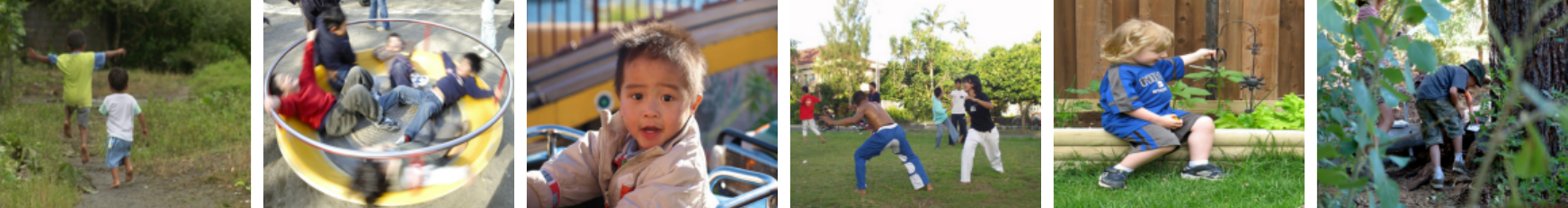}
        \caption{$\tilde{\mP}_{\text{a child playing outdoors}}(\mI - \tilde{\mP}_{\text{near a road}})$}
    \end{subfigure}

     \begin{subfigure}{\linewidth}
        \centering
        \includegraphics[width=\linewidth]{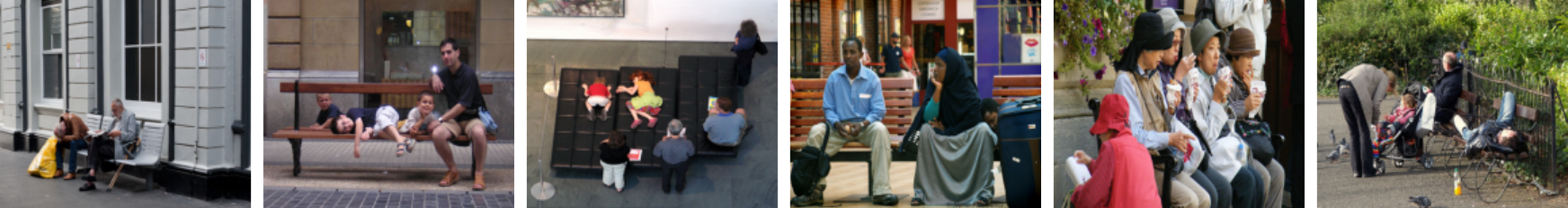}
        \caption{$\tilde{\mP}_\text{people} \tilde{\mP}_\text{sitting} \tilde{\mP}_\text{on a bench}$}
    \end{subfigure}
    
    \caption{Flickr30k retrieval from composition of natural language queries via subspace intersections and complements. }
    \label{fig:flickr30k_retrieval}
\end{figure}

\paragraph{Emergent compositionality.} As evidenced by the results in Table \ref{tab:compositional_entailment}, a key advantage of subspace embeddings is their emergent compositionality, which arises from the geometry of the embeddings without explicit training signals. Fig.~\ref{fig:flickr30k_retrieval} provides an example illustrating this inherent compositionality for conjunctions $\tilde{\mP}_i\tilde{\mP}_j$ and negations $\mI-\tilde{\mP}$ of queries, in a retrieval setting. For a query formed by a logical composition of concept subspaces, we retrieve images from Flickr30k \citep{young2014image} whose caption subspaces have the largest $\mathrm{NIS}$($\tilde{\mP}_\text{query}$ $\mid$ $\tilde{\mP}_\text{caption}$). Each caption subspace is computed with our mpnet-base-v2 + SPH (SE$^{128}$) model, fine-tuned on SNLI. The results demonstrate that subspaces enable compositional retrieval, allowing for the search of novel concepts or query editing through geometric operations. 

\section{Limitations}
\label{sec:limitations}
In this section, we discuss the boundaries of our framework and identify promising directions for future research.

\paragraph{Approximate logic.} Our framework uses soft projectors with eigenvalues in $[0,1)$ to ensure end-to-end differentiability.  Consequently, the induced logical operations are algebraic approximations rather than truly symbolic. While this prevents the execution of discrete Boolean logic, our results on SNLI suggest that it allows the model to capture graded entailment and semantic ambiguity. Future work may explore annealing the hyperparameter $\lambda \to 0$ to recover hard logic for tasks requiring strict symbolic consistency.

\paragraph{Storage efficiency.} Naively storing the full $d\times d$ operators would incur a quadratic memory cost. However, our analysis of the \emph{effective rank} (Fig. \ref{fig:dim_x_deg}) reveals that the learned subspaces are intrinsically low-dimensional. This indicates that the information content of our representations scales with $\mathcal{O}(d \cdot k)$ rather than $\mathcal{O}(d^2)$, where $k$ is the average subspace rank. Future implementations can explicitly enforce low-rank constraints to match the storage footprint of standard vector embeddings without loss of expressivity.

\paragraph{Compositional entailment dataset.} As with any LLM-elicited benchmark, the data may inherit distributional biases from the generating model. Manual validation and baseline filtering mitigate this for the specific composition operations under study, but we do not claim the set is representative of compositional reasoning in natural language more broadly.
\section{Conclusion}
\label{sec:conclusion}
This paper proposed \emph{subspace embeddings}, a geometric framework that overcomes the intrinsic inability of point vectors to model hierarchical and asymmetric relationships. By representing concepts as linear subspaces, we introduce an inductive bias where generality corresponds to dimensionality and entailment to geometric inclusion. As shown by our empirical evaluation, subspace embeddings achieve state-of-the-art performance on hierarchical modeling and natural language inference, while enabling zero-shot logical composition, namely negation, where standard vector representations degrade to random performance. Crucially, this expressivity does not come at the cost of scalability, as our formulation preserves compatibility with standard Maximum Inner Product Search pipelines, and the data-dependent subspace ranks allow for considerable dimensionality reduction. By reconciling the efficiency of vector retrieval with the hierarchical and compositional structure of natural language, our work offers a promising path toward more expressive and interpretable representation learning systems.

\begin{acks}
    This work is funded by LARSyS funding (LA/P/0083/2020: DOI {\small\nolinkurl{10.54499/LA/P/0083/2020}}, UID/50009/2025: DOI {\small\nolinkurl{10.54499/UID/50009/2025}}), through Fundação para a Ciência e a Tecnologia. G. Moreira is also supported via grant {\small\nolinkurl{SFRH/BD/151466/2021}} through the Carnegie Mellon Portugal program. J. P. Costeira and M. Marques are also supported by the PT Smart Retail project (PRR - {\small\nolinkurl{02/C05-i11/2024.C645440011-00000062}}), through IAPMEI - Agência para a Competitividade e Inovação. 
\end{acks}

\bibliographystyle{ACM-Reference-Format}
\balance
\bibliography{references}

@String{Computer = "{IEEE} Computer" }

@String{Springer = "Springer-Verlag" }

@inproceedings{zhang2021cone,
  title={Cone: Cone embeddings for multi-hop reasoning over knowledge graphs},
  author={Zhang, Zhanqiu and Wang, Jie and Chen, Jiajun and Ji, Shuiwang and Wu, Feng},
  journal={Advances in Neural Information Processing Systems},
  volume={34},
  pages={19172--19183},
  year={2021}
}

@inproceedings{ganea2018hyperbolic_cones,
  title={Hyperbolic entailment cones for learning hierarchical embeddings},
  author={Ganea, Octavian and B{\'e}cigneul, Gary and Hofmann, Thomas},
  booktitle={International conference on machine learning},
  pages={1646--1655},
  year={2018},
  organization={PMLR}
}

@inproceedings{ganea2018hyperbolic,
  title={Hyperbolic neural networks},
  author={Ganea, Octavian and B{\'e}cigneul, Gary and Hofmann, Thomas},
  journal={Advances in neural information processing systems},
  volume={31},
  year={2018}
}

@inproceedings{sala2018representation,
  title={Representation tradeoffs for hyperbolic embeddings},
  author={Sala, Frederic and De Sa, Chris and Gu, Albert and R{\'e}, Christopher},
  booktitle={International conference on machine learning},
  pages={4460--4469},
  year={2018},
  organization={PMLR}
}

@inproceedings{nickel2018learning,
  title={Learning continuous hierarchies in the lorentz model of hyperbolic geometry},
  author={Nickel, Maximillian and Kiela, Douwe},
  booktitle={International conference on machine learning},
  pages={3779--3788},
  year={2018},
  organization={PMLR}
}

@inproceedings{nickel2017poincare,
  title={Poincar{\'e} embeddings for learning hierarchical representations},
  author={Nickel, Maximillian and Kiela, Douwe},
  journal={Advances in neural information processing systems},
  volume={30},
  year={2017}
}

@inproceedings{moreira2024hyperbolic,
  title={Hyperbolic vs Euclidean embeddings in few-shot learning: Two sides of the same coin},
  author={Moreira, Gabriel and Marques, Manuel and Costeira, Jo{\~a}o Paulo and Hauptmann, Alexander},
  booktitle={Proceedings of the IEEE/CVF Winter Conference on Applications of Computer Vision},
  pages={2082--2090},
  year={2024}
}

@inproceedings{bai2021modeling,
  title={Modeling heterogeneous hierarchies with relation-specific hyperbolic cones},
  author={Bai, Yushi and Ying, Zhitao and Ren, Hongyu and Leskovec, Jure},
  journal={Advances in Neural Information Processing Systems},
  volume={34},
  pages={12316--12327},
  year={2021}
}

@inproceedings{he2024language,
  title={Language models as hierarchy encoders},
  author={He, Yuan and Yuan, Moy and Chen, Jiaoyan and Horrocks, Ian},
  journal={Advances in Neural Information Processing Systems},
  volume={37},
  pages={14690--14711},
  year={2024}
}

@inproceedings{li2018smoothing,
  title={Smoothing the geometry of probabilistic box embeddings},
  author={Li, Xiang and Vilnis, Luke and Zhang, Dongxu and Boratko, Michael and McCallum, Andrew},
  booktitle={International Conference on Learning Representations},
  year={2018}
}

@inproceedings{ren2020query2box,
  title={Query2box: Reasoning Over Knowledge Graphs In Vector Space Using Box Embeddings},
  author={Ren, H and Hu, W and Leskovec, J},
  booktitle={International Conference on Learning Representations (ICLR)},
  year={2020}
}

@inproceedings{ren2020beta,
  title={Beta embeddings for multi-hop logical reasoning in knowledge graphs},
  author={Ren, Hongyu and Leskovec, Jure},
  journal={Advances in Neural Information Processing Systems},
  volume={33},
  pages={19716--19726},
  year={2020}
}

@inproceedings{vilnis2018probabilistic,
  title={Probabilistic Embedding of Knowledge Graphs with Box Lattice Measures},
  author={Vilnis, Luke and Li, Xiang and Murty, Shikhar and Mccallum, Andrew},
  booktitle={Proceedings of the 56th Annual Meeting of the Association for Computational Linguistics (Volume 1: Long Papers)},
  pages={263--272},
  year={2018}
}

@misc{vilnis2014word,
      title={Word Representations via Gaussian Embedding}, 
      author={Luke Vilnis and Andrew McCallum},
      year={2015},
      eprint={1412.6623},
      archivePrefix={arXiv},
      primaryClass={cs.CL},
      url={https://arxiv.org/abs/1412.6623}, 
}

@misc{park2024geometry,
      title={The Geometry of Categorical and Hierarchical Concepts in Large Language Models}, 
      author={Kiho Park and Yo Joong Choe and Yibo Jiang and Victor Veitch},
      year={2025},
      eprint={2406.01506},
      archivePrefix={arXiv},
      primaryClass={cs.CL},
      url={https://arxiv.org/abs/2406.01506}, 
}

@inproceedings{weller2023nevir,
  title={NevIR: Negation in Neural Information Retrieval},
  author={Weller, Orion and Lawrie, Dawn and Van Durme, Benjamin},
  booktitle={Proceedings of the 18th Conference of the European Chapter of the Association for Computational Linguistics (Volume 1: Long Papers)},
  pages={2274--2287},
  year={2024}
}

@misc{singh2024learn,
      title={Learn "No" to Say "Yes" Better: Improving Vision-Language Models via Negations}, 
      author={Jaisidh Singh and Ishaan Shrivastava and Mayank Vatsa and Richa Singh and Aparna Bharati},
      year={2024},
      eprint={2403.20312},
      archivePrefix={arXiv},
      primaryClass={cs.CV},
      url={https://arxiv.org/abs/2403.20312}, 
}

@inproceedings{quantmeyer2024and,
  title={How and where does CLIP process negation?},
  author={Quantmeyer, Vincent and Mosteiro, Pablo and Gatt, Albert},
  booktitle={Proceedings of the 3rd Workshop on Advances in Language and Vision Research (ALVR)},
  pages={59--72},
  year={2024}
}

@misc{yuksekgonul2022and,
      title={When and why vision-language models behave like bags-of-words, and what to do about it?}, 
      author={Mert Yuksekgonul and Federico Bianchi and Pratyusha Kalluri and Dan Jurafsky and James Zou},
      year={2023},
      eprint={2210.01936},
      archivePrefix={arXiv},
      primaryClass={cs.CV},
      url={https://arxiv.org/abs/2210.01936}, 
}

@inproceedings{moreira2025learning,
  title={Learning Visual-Semantic Subspace Representations},
  author={Moreira, Gabriel and Hauptmann, Alexander and Marques, Manuel and Costeira, Jo{\~a}o Paulo},
  journal={The 28th International Conference on Artificial Intelligence and Statistics},
  year={2025}
}

@article{miller1995wordnet,
  title={WordNet: a lexical database for English},
  author={Miller, George A},
  journal={Communications of the ACM},
  volume={38},
  number={11},
  pages={39--41},
  year={1995},
  publisher={ACM New York, NY, USA}
}

@inproceedings{suzuki2019hyperbolic,
  title={Hyperbolic disk embeddings for directed acyclic graphs},
  author={Suzuki, Ryota and Takahama, Ryusuke and Onoda, Shun},
  booktitle={International Conference on Machine Learning},
  pages={6066--6075},
  year={2019},
  organization={PMLR}
}

@inproceedings{xiong2022hyperbolic,
  title={Hyperbolic embedding inference for structured multi-label prediction},
  author={Xiong, Bo and Cochez, Michael and Nayyeri, Mojtaba and Staab, Steffen},
  journal={Advances in Neural Information Processing Systems},
  volume={35},
  pages={33016--33028},
  year={2022}
}

@inproceedings{dhall2020hierarchical,
  title={Hierarchical image classification using entailment cone embeddings},
  author={Dhall, Ankit and Makarova, Anastasia and Ganea, Octavian and Pavllo, Dario and Greeff, Michael and Krause, Andreas},
  booktitle={Proceedings of the IEEE/CVF conference on computer vision and pattern recognition workshops},
  pages={836--837},
  year={2020}
}

@inproceedings{poppi2025hyperbolic,
  title={Hyperbolic Safety-Aware Vision-Language Models},
  author={Poppi, Tobia and Kasarla, Tejaswi and Mettes, Pascal and Baraldi, Lorenzo and Cucchiara, Rita},
  booktitle={Proceedings of the Computer Vision and Pattern Recognition Conference},
  pages={4222--4232},
  year={2025}
}

@misc{vendrov2015order,
      title={Order-Embeddings of Images and Language}, 
      author={Ivan Vendrov and Ryan Kiros and Sanja Fidler and Raquel Urtasun},
      year={2016},
      eprint={1511.06361},
      archivePrefix={arXiv},
      primaryClass={cs.LG},
      url={https://arxiv.org/abs/1511.06361}, 
}

@article{johnson2019billion,
  title={Billion-scale similarity search with {GPUs}},
  author={Johnson, Jeff and Douze, Matthijs and J{\'e}gou, Herv{\'e}},
  journal={IEEE Transactions on Big Data},
  volume={7},
  number={3},
  pages={535--547},
  year={2019},
  publisher={IEEE}
}

@misc{douze2024faiss,
      title={The Faiss library}, 
      author={Matthijs Douze and Alexandr Guzhva and Chengqi Deng and Jeff Johnson and Gergely Szilvasy and Pierre-Emmanuel Mazaré and Maria Lomeli and Lucas Hosseini and Hervé Jégou},
      year={2025},
      eprint={2401.08281},
      archivePrefix={arXiv},
      primaryClass={cs.LG},
      url={https://arxiv.org/abs/2401.08281}, 
}

@inproceedings{radford2021learning,
  title={Learning transferable visual models from natural language supervision},
  author={Radford, Alec and Kim, Jong Wook and Hallacy, Chris and Ramesh, Aditya and Goh, Gabriel and Agarwal, Sandhini and Sastry, Girish and Askell, Amanda and Mishkin, Pamela and Clark, Jack and others},
  booktitle={International conference on machine learning},
  pages={8748--8763},
  year={2021},
  organization={PmLR}
}

@inproceedings{mikolov2013distributed,
  title={Distributed representations of words and phrases and their compositionality},
  author={Mikolov, Tomas and Sutskever, Ilya and Chen, Kai and Corrado, Greg S and Dean, Jeff},
  journal={Advances in neural information processing systems},
  volume={26},
  year={2013}
}

@book{ganter2024formal,
  title={Formal concept analysis: mathematical foundations},
  author={Ganter, Bernhard and Wille, Rudolf},
  year={2024},
  publisher={Springer Nature}
}

@inproceedings{da2009normalized,
  title={The normalized subspace inclusion: Robust clustering of motion subspaces},
  author={Da Silva, Nuno Pinho and Costeira, Joao Paulo},
  booktitle={2009 IEEE 12th International Conference on Computer Vision},
  pages={1444--1450},
  year={2009},
  organization={IEEE}
}

@inproceedings{athiwaratkun2018hierarchical,
  title={Hierarchical density order embeddings},
  author={Athiwaratkun, Ben and Wilson, Andrew Gordon},
  booktitle={6th International Conference on Learning Representations, ICLR 2018},
  year={2018}
}

@inproceedings{yushadow,
  title={Shadow Cones: A Generalized Framework for Partial Order Embeddings},
  author={Yu, Tao and Liu, Toni JB and Tseng, Albert and De Sa, Christopher},
  booktitle={The Twelfth International Conference on Learning Representations},
  year={2024}
}

@book{van2004geometry,
  title={The geometry of information retrieval},
  author={Van Rijsbergen, Cornelis Joost},
  year={2004},
  publisher={Cambridge University Press}
}

@inproceedings{devlin2019bert,
  title={Bert: Pre-training of deep bidirectional transformers for language understanding},
  author={Devlin, Jacob and Chang, Ming-Wei and Lee, Kenton and Toutanova, Kristina},
  booktitle={Proceedings of the 2019 conference of the North American chapter of the association for computational linguistics: human language technologies, volume 1 (long and short papers)},
  pages={4171--4186},
  year={2019}
}

@inproceedings{reimers2019sentence,
  title={Sentence-BERT: Sentence Embeddings using Siamese BERT-Networks},
  author={Reimers, Nils and Gurevych, Iryna},
  booktitle={Proceedings of the 2019 Conference on Empirical Methods in Natural Language Processing and the 9th International Joint Conference on Natural Language Processing (EMNLP-IJCNLP)},
  pages={3982--3992},
  year={2019}
}

@article{lewis2020retrieval,
  title={Retrieval-augmented generation for knowledge-intensive nlp tasks},
  author={Lewis, Patrick and Perez, Ethan and Piktus, Aleksandra and Petroni, Fabio and Karpukhin, Vladimir and Goyal, Naman and K{\"u}ttler, Heinrich and Lewis, Mike and Yih, Wen-tau and Rockt{\"a}schel, Tim and others},
  journal={Advances in neural information processing systems},
  volume={33},
  pages={9459--9474},
  year={2020}
}

@inproceedings{li2022blip,
  title={Blip: Bootstrapping language-image pre-training for unified vision-language understanding and generation},
  author={Li, Junnan and Li, Dongxu and Xiong, Caiming and Hoi, Steven},
  booktitle={International conference on machine learning},
  pages={12888--12900},
  year={2022},
  organization={PMLR}
}

@inproceedings{gokhale2020vqa,
  title={Vqa-lol: Visual question answering under the lens of logic},
  author={Gokhale, Tejas and Banerjee, Pratyay and Baral, Chitta and Yang, Yezhou},
  booktitle={European conference on computer vision},
  pages={379--396},
  year={2020},
  organization={Springer}
}

@misc{zhang2025negvqavisionlanguagemodels,
      title={NegVQA: Can Vision Language Models Understand Negation?}, 
      author={Yuhui Zhang and Yuchang Su and Yiming Liu and Serena Yeung-Levy},
      year={2025},
      eprint={2505.22946},
      archivePrefix={arXiv},
      primaryClass={cs.CL},
      url={https://arxiv.org/abs/2505.22946}, 
}

@inproceedings{alhamoud2025vision,
  title={Vision-language models do not understand negation},
  author={Alhamoud, Kumail and Alshammari, Shaden and Tian, Yonglong and Li, Guohao and Torr, Philip HS and Kim, Yoon and Ghassemi, Marzyeh},
  booktitle={Proceedings of the Computer Vision and Pattern Recognition Conference},
  pages={29612--29622},
  year={2025}
}

@misc{kingma2017adammethodstochasticoptimization,
      title={Adam: A Method for Stochastic Optimization}, 
      author={Diederik P. Kingma and Jimmy Ba},
      year={2017},
      eprint={1412.6980},
      archivePrefix={arXiv},
      primaryClass={cs.LG},
      url={https://arxiv.org/abs/1412.6980}, 
}

@misc{oord2019representationlearningcontrastivepredictive,
      title={Representation Learning with Contrastive Predictive Coding}, 
      author={Aaron van den Oord and Yazhe Li and Oriol Vinyals},
      year={2019},
      eprint={1807.03748},
      archivePrefix={arXiv},
      primaryClass={cs.LG},
      url={https://arxiv.org/abs/1807.03748}, 
}

@inproceedings{bowman2015large,
  title={A large annotated corpus for learning natural language inference},
  author={Bowman, Samuel and Angeli, Gabor and Potts, Christopher and Manning, Christopher D},
  booktitle={Proceedings of the 2015 Conference on Empirical Methods in Natural Language Processing},
  pages={632--642},
  year={2015}
}

@inproceedings{srivastava2020inductive,
  title={Inductive quantum embedding},
  author={Srivastava, Santosh Kumar and Khandelwal, Dinesh and Madan, Dhiraj and Garg, Dinesh and Karanam, Hima and Subramaniam, L Venkata},
  booktitle={Advances in Neural Information Processing Systems},
  volume={33},
  pages={16012--16024},
  year={2020}
}

@inproceedings{garg2019quantum,
  title={Quantum embedding of knowledge for reasoning},
  author={Garg, Dinesh and Ikbal, Shajith and Srivastava, Santosh K and Vishwakarma, Harit and Karanam, Hima and Subramaniam, L Venkata},
  booktitle={Advances in Neural Information Processing Systems},
  volume={32},
  year={2019}
}

@article{young2014image,
  title={From image descriptions to visual denotations: New similarity metrics for semantic inference over event descriptions},
  author={Young, Peter and Lai, Alice and Hodosh, Micah and Hockenmaier, Julia},
  journal={Transactions of the Association for Computational Linguistics},
  volume={2},
  pages={67--78},
  year={2014},
  publisher={MIT Press}
}

@inproceedings{lee2019set,
  title={Set transformer: A framework for attention-based permutation-invariant neural networks},
  author={Lee, Juho and Lee, Yoonho and Kim, Jungtaek and Kosiorek, Adam and Choi, Seungjin and Teh, Yee Whye},
  booktitle={International conference on machine learning},
  pages={3744--3753},
  year={2019},
  organization={PMLR}
}

@article{behnamghader2024llm2vec,
  title={Llm2vec: Large language models are secretly powerful text encoders},
  author={BehnamGhader, Parishad and Adlakha, Vaibhav and Mosbach, Marius and Bahdanau, Dzmitry and Chapados, Nicolas and Reddy, Siva},
  journal={arXiv preprint arXiv:2404.05961},
  year={2024}
}

@article{oquab2023dinov2,
  title={Dinov2: Learning robust visual features without supervision},
  author={Oquab, Maxime and Darcet, Timoth{\'e}e and Moutakanni, Th{\'e}o and Vo, Huy and Szafraniec, Marc and Khalidov, Vasil and Fernandez, Pierre and Haziza, Daniel and Massa, Francisco and El-Nouby, Alaaeldin and others},
  journal={arXiv preprint arXiv:2304.07193},
  year={2023}
}

@article{kusupati2022matryoshka,
  title={Matryoshka representation learning},
  author={Kusupati, Aditya and Bhatt, Gantavya and Rege, Aniket and Wallingford, Matthew and Sinha, Aditya and Ramanujan, Vivek and Howard-Snyder, William and Chen, Kaifeng and Kakade, Sham and Jain, Prateek and others},
  journal={Advances in Neural Information Processing Systems},
  volume={35},
  pages={30233--30249},
  year={2022}
}

@article{wang2024multilingual,
  title={Multilingual e5 text embeddings: A technical report},
  author={Wang, Liang and Yang, Nan and Huang, Xiaolong and Yang, Linjun and Majumder, Rangan and Wei, Furu},
  journal={arXiv preprint arXiv:2402.05672},
  year={2024}
}

@inproceedings{gyurek2024binder,
  title={Binder: Hierarchical concept representation through order embedding of binary vectors},
  author={Gyurek, Croix and Talukder, Niloy and Hasan, Mohammad Al},
  booktitle={Proceedings of the 30th ACM SIGKDD Conference on Knowledge Discovery and Data Mining},
  pages={980--991},
  year={2024}
}

\appendix
\section{Error Bounds for Soft Projectors}
\label{app:sec:theoretical_bounds}
\begin{lemma}
    \label{lemma:evd_soft_projector}
    Let $\mX=\mU\mathbf{\Sigma}\mV^\top$, where $\mU\in O(d)$, and define $\tilde{\mP} := \mX(\mX^\top \mX + \lambda\mI)^{-1} \mX^\top$. We have
    \begin{equation}
         \tilde{\mP} = \mU\mathbf{\Sigma}^2(\mathbf{\Sigma}^2 + \lambda\mI)^{-1}\mU^\top
    \end{equation}
    and the spectrum of $\tilde{\mP}$ is $\left\{\frac{\sigma_i^2}{\sigma_i^2+\lambda}\right\}_{i=1}^d$.
    \begin{proof}
        Letting the SVD of $\mX$ be $\mU\mathbf{\Sigma}\mV^\top$,
        \begin{align}
           \tilde{\mP} &= \mU\mathbf{\Sigma}\mV^\top(\mV\mathbf{\Sigma}^2\mV^\top + \lambda\mI)^{-1} \mV\mathbf{\Sigma}\mU^\top 
           = \mU\mathbf{\Sigma}^2(\mathbf{\Sigma}^2 + \lambda\mI)^{-1}\mU^\top.
        \end{align}
        The spectrum of $\tilde{\mP}$ is the diagonal of $\mathbf{\Sigma}^2(\mathbf{\Sigma}^2 + \lambda\mI)^{-1}$. 
    \end{proof}
\end{lemma}

\begin{proposition}[Frobenius norm error]
    \label{proposition:frobenius_error}
    Let $\mX=\mU\mathbf{\Sigma}\mV^\top$ be rank-$r$, where $\mU\in O(d)$ and let the orthogonal projector onto $\mathrm{Span}(\mX)$ be $\mP=\mU\mJ_r\mU^\top$. Define $\tilde{\mP} := \mX(\mX^\top \mX + \lambda\mI)^{-1} \mX^\top$. Then,
    \begin{equation}
        \| \mP - \tilde{\mP} \|_F \leq \frac{\lambda\sqrt{r}}{\sigma_r^2+\lambda}.
    \end{equation}
    \begin{proof}
        Let $\mJ_r=\mathrm{BlockDiag}(\mI_r, \mathbf{0}_{d-r})$, where $r = \mathrm{rank}(\mX)$ and write the SVD of the orthogonal projector as $\mP=\mU \mJ_r \mU^\top$ for $\mU\in O(d)$. Using Lemma \ref{lemma:evd_soft_projector}, we can write $\mP - \tilde{\mP} = \mU(\mJ_r - \mathbf{\Sigma}^2(\mathbf{\Sigma}^2 + \lambda\mI)^{-1})\mU^\top$. The Frobenius norm is invariant to orthogonal transformations $\mU$, hence
        \begin{align}
            \| \mP - \tilde{\mP} \|_F^2 &= \| \mJ_r - \mathbf{\Sigma}^2(\mathbf{\Sigma}^2 + \lambda\mI)^{-1} \|_F^2 \nonumber \\ &= \sum_{i=1}^r\left(1 - \frac{\sigma_i^2}{\sigma_i^2+\lambda}\right)^2 \leq r\left(\frac{\lambda}{\sigma_r^2+\lambda}\right)^2.
        \end{align}
        Therefore, $\| \mP - \tilde{\mP} \|_F \leq \frac{\lambda\sqrt{r}}{\sigma_r^2+\lambda}$.
    \end{proof}
\end{proposition}

\begin{proposition}[Operator norm error]
    \label{proposition:operator_norm_error}
    Let $\mX=\mU\mathbf{\Sigma}\mV^\top$ be rank-$r$, where $\mU\in O(d)$, and let the orthogonal projector onto $\mathrm{Span}(\mX)$ be $\mP=\mU\mJ_r\mU^\top$. Define $\tilde{\mP} := \mX(\mX^\top \mX + \lambda\mI)^{-1} \mX^\top$. Then,
    \begin{equation}
        \| \mP - \tilde{\mP} \|_2 = \frac{\lambda}{\sigma_r^2+\lambda}.
    \end{equation}
    \begin{proof}
        Let $\mJ_r=\mathrm{BlockDiag}(\mI_r, \mathbf{0}_{d-r})$, where $r = \mathrm{rank}(\mX)$ and write the SVD of the orthogonal projector as $\mP=\mU \mJ_r \mU^\top$ for $\mU\in O(d)$. Using Lemma \ref{lemma:evd_soft_projector}, we can write $\mP - \tilde{\mP} = \mU(\mJ_r - \mathbf{\Sigma}^2(\mathbf{\Sigma}^2 + \lambda\mI)^{-1})\mU^\top$. The operator norm is invariant to orthogonal transformations $\mU$, hence
        \begin{align}
            \| \mP - \tilde{\mP} \|_2 &= \| \mJ_r - \mathbf{\Sigma}^2(\mathbf{\Sigma}^2 + \lambda\mI)^{-1} \|_2 \nonumber \\ &= \max\left\{1 - \frac{\sigma_i^2}{\sigma_i^2+\lambda}\right\}_{i=1}^r = \frac{\lambda}{\sigma_r^2+\lambda}.
        \end{align}
        Therefore, $\| \mP - \tilde{\mP} \|_2 = \frac{\lambda}{\sigma_r^2+\lambda}$.
    \end{proof}
\end{proposition}

\begin{corollary}[Negation operator error]
    Let $\mX$, $\mP$ and $\tilde{\mP}$ be in the conditions of Proposition \ref{proposition:frobenius_error}. Then,
    \begin{equation}
        \| (\mI-\mP) - (\mI-\tilde{\mP}) \|_2 =  \frac{\lambda}{\sigma_r^2+\lambda}.
    \end{equation}
    \begin{proof}
        Note that $\| (\mI-\mP) - (\mI-\tilde{\mP}) \|_2 = \| \mP -\tilde{\mP} \|_2$ and apply Proposition \ref{proposition:operator_norm_error}.      
    \end{proof}
\end{corollary}

\begin{proposition}[Trace error]
    \label{proposition:trace_bounds}
    Let $\mX=\mU\mathbf{\Sigma}\mV^\top$ be rank-$r$, where $\mU\in O(d)$, and let the orthogonal projector onto $\mathrm{Span}(\mX)$ be $\mP=\mU\mJ_r\mU^\top$. Define $\tilde{\mP} := \mX(\mX^\top \mX + \lambda\mI)^{-1} \mX^\top$. Then,
    \begin{equation}
        \left|\mathrm{Tr}(\mP) - \mathrm{Tr}(\tilde{\mP})\right| \leq \frac{\lambda r}{\sigma_r^2+\lambda} 
    \end{equation}
    \begin{proof}
        Write $\mP - \tilde{\mP} = \mU(\mJ_r - \mathbf{\Sigma}^2(\mathbf{\Sigma}^2 + \lambda\mI)^{-1})\mU^\top$. Then,
        \begin{align}
            \left|\mathrm{Tr}(\mP) - \mathrm{Tr}(\tilde{\mP})\right| &=\left|\mathrm{Tr}(\mJ_r - \mathbf{\Sigma}^2(\mathbf{\Sigma}^2 + \lambda\mI)^{-1})\right| 
            \nonumber \\ &= \sum_{i=1}^r \left(1 - \frac{\sigma_i^2}{\sigma_i^2+\lambda}\right) 
            \leq \frac{\lambda r}{\sigma_r^2+\lambda} \qedhere 
        \end{align}
    \end{proof}
\end{proposition}

\begin{corollary}[Subspace rank error]
    Letting $r:=\mathrm{rank}(\mX)$, the relative error of estimating $r$ via $\mathrm{Tr}(\tilde{\mP})$ satisfies
    \begin{equation}
        \frac{\left|\mathrm{Tr}(\tilde{\mP}) - r\right|}{r} \leq \frac{\lambda}{\sigma_r^2+\lambda}.
    \end{equation}
    \begin{proof}
        Note that $r=\mathrm{Tr}(\mP)$ and use Proposition \ref{proposition:trace_bounds}.
    \end{proof}
\end{corollary}

\begin{proposition}[Subspace similarity error]
    Let $\mX_i$ and $\mX_j$ be rank-$r_i$ and $r_j$ matrices, with singular values $\{\sigma_k\}_{k=1}^d$ and $\{\eta_k\}_{k=1}^d$ (in descending order), respectively. Denote by $\mP_i, \tilde{\mP}_i$ and $\mP_j, \tilde{\mP}_j$ the respective orthogonal and soft projectors. Then,
    \begin{equation}
        \left|\mathrm{Tr}(\mP_i\mP_j) - \mathrm{Tr}(\tilde{\mP}_i\tilde{\mP}_j)\right| \leq \sqrt{r_i r_j}\left(\frac{\lambda}{\sigma_r^2+\lambda} + \frac{\lambda}{\eta_r^2 + \lambda}\frac{\sigma_r^2}{\sigma_r^2+\lambda} \right)
    \end{equation}
    \begin{proof}
        We have
        \begin{align}
            \left|\mathrm{Tr}(\mP_i\mP_j) - \mathrm{Tr}(\tilde{\mP}_i\tilde{\mP}_j)\right| &= \left| \mathrm{Tr}((\mP_i - \tilde{\mP}_i)\mP_j) + \mathrm{Tr}((\mP_j - \tilde{\mP}_j)\tilde{\mP}_i)\right| \nonumber \\
            &\leq \left| \mathrm{Tr}((\mP_i - \tilde{\mP}_i)\mP_j)  \right| + \left| \mathrm{Tr}((\mP_j - \tilde{\mP}_j)\tilde{\mP}_i)\right|.
        \end{align}
        Apply Cauchy-Schwartz to both terms, we arrive at
        \begin{align}
             \left|\mathrm{Tr}(\mP_i\mP_j) - \mathrm{Tr}(\tilde{\mP}_i\tilde{\mP}_j)\right| &\leq \|\mP_i-\tilde{\mP}_i\|_F\|\mP_j\|_F + \|\mP_j - \tilde{\mP}_j\|_F \|\tilde{\mP}_i\|_F \nonumber \\
        \end{align}
        and we can replace $\sqrt{r_j} = \|\mP_j\|_F$, $\sqrt{r_i} = \|\mP_i\|_F$ and employ Proposition \ref{proposition:frobenius_error},
        \begin{align}
            \left|\mathrm{Tr}(\mP_i\mP_j) - \mathrm{Tr}(\tilde{\mP}_i\tilde{\mP}_j)\right| &\leq \|\mP_i-\tilde{\mP}_i\|_F\sqrt{r_j} + \|\mP_j - \tilde{\mP}_j\|_F \|\tilde{\mP}_i\|_F  \nonumber \\
            &\leq \sqrt{r_ir_j} \frac{\lambda}{\sigma_r^2+\lambda} +\sqrt{r_j} \frac{\lambda}{\eta_r^2+\lambda}\|\tilde{\mP}_i\|_F.
        \end{align}
        Finally, note that $\|\tilde{\mP}_i\|_F \leq \sqrt{r_i}$
        \begin{equation}
            \left|\mathrm{Tr}(\mP_i\mP_j) - \mathrm{Tr}(\tilde{\mP}_i\tilde{\mP}_j)\right|  \leq \sqrt{r_i r_j}\left(\frac{\lambda}{\sigma_r^2+\lambda} + \frac{\lambda}{\eta_r^2 + \lambda} \right).\qedhere
        \end{equation}
    \end{proof}
\end{proposition}

\begin{proposition}[Intersection operator error]
    Let $\mX_i$ and $\mX_j$ be rank-$r_i$ and $r_j$ matrices, with singular values $\{\sigma_k\}_{k=1}^d$ and $\{\eta_k\}_{k=1}^d$ (in descending order), respectively. Denote by $\mP_i, \tilde{\mP}_i$ and $\mP_j, \tilde{\mP}_j$ the respective orthogonal and soft projectors. Then,
    \begin{equation}
        \|\mP_i\mP_j - \tilde{\mP}_i\tilde{\mP}_j\|_2 \leq \frac{\lambda}{\sigma_r^2 + \lambda} + \frac{\lambda}{\eta_r^2 + \lambda}.
    \end{equation}
    \begin{proof}
        From writing $\mP_i\mP_j - \tilde{\mP}_i\tilde{\mP}_j = (\mP_i - \tilde{\mP}_i)\mP_j + (\mP_j - \tilde{\mP}_j)\tilde{\mP}_i$ and applying the triangle inequality
        \begin{align}
            \|\mP_i\mP_j - \tilde{\mP}_i\tilde{\mP}_j\|_2 &= \|(\mP_i - \tilde{\mP}_i)\mP_j + (\mP_j - \tilde{\mP}_j)\tilde{\mP}_i \|_2 \nonumber \\
            &\leq \|\mP_i - \tilde{\mP}_i \|_2 \|\mP_j\|_2 + \| \mP_j - \tilde{\mP}_j\|_2 \|\tilde{\mP}_i \|_2.
        \end{align}
        Noting that $ \|\mP_j\|_2 \leq 1$ and $\|\tilde{\mP}_i \|_2 \leq 1$ and using Proposition \ref{proposition:operator_norm_error}, we have
        \begin{equation}
            \|\mP_i\mP_j - \tilde{\mP}_i\tilde{\mP}_j\|_2 \leq \frac{\lambda}{\sigma_r^2 + \lambda} + \frac{\lambda}{\eta_r^2 + \lambda}.\;\qedhere
        \end{equation}
    \end{proof}
\end{proposition}

\section{Gradients of Soft Projection Matrices}
\label{app:sec:training_dynamics}
The gradient of $\mathrm{Tr}(\tilde{\mP}_i \tilde{\mP}_j)$ with respect to $\mX_i$ can be derived from the identity
\begin{align}
    &\nabla_{\mX}\mathrm{Tr}\left((\mA+\mX^\top \mC\mX)^{-1} (\mX^\top \mB\mX)\right) = \nonumber \\
    &-2\mC\mX(\mA+\mX^\top \mC\mX)^{-1} \mX^\top \mB\mX (\mA+\mX^\top \mC\mX)^{-1} \nonumber \\ &+ 2 \mB\mX(\mA + \mX^\top \mC \mX)^{-1}.
\end{align}
We have then
\begin{align}
    \nabla_{\mX_i} \mathrm{Tr}\left(\tilde{\mP}_i \tilde{\mP}_j\right) &= 2(\mI - \tilde{\mP}_i)\tilde{\mP}_j \mX_i(\mX_i^\top \mX_i + \lambda \mI)^{-1} \nonumber \\
    &\propto\underbrace{\tilde{\mP}_i^\perp\tilde{\mP}_j}_{\text{New information}} \underbrace{\mX_i(\mX_i^\top \mX_i + \lambda \mI)^{-1}}_{\text{Spectral scaling}}.
\end{align}
The spectral scaling factor $\mX_i(\mX_i^\top \mX_i + \lambda \mI)^{-1}$ acts as a low-pass filter on $\mX_i$. If we write the SVD of $\mX_i$ as $\mX_i = \mU\mathbf{\Sigma}\mV^\top$, then $\mX_i(\mX_i^\top \mX_i + \lambda \mI)^{-1} = \mU \mathbf{\Sigma}(\mathbf{\Sigma}^2 + \lambda \mI)^{-1} \mV^\top$. As a result, high-energy directions (associated with large singular values) are attenuated, while low-energy directions are amplified. This ensures that updates to $\mX_i$ preserve dominant, well-supported directions while adapting underrepresented ones.

The component $\tilde{\mP}_i^\perp\tilde{\mP}_j$, where $\tilde{\mP}_i^\perp = \mI - \tilde{\mP}_i$, indicates that gradient flow occurs only along directions present in subspace $j$ but orthogonal to subspace $i$, formally, in $\mathrm{range}(\tilde{\mP}_j)\cap\mathrm{null}(\tilde{\mP}_i)$. Thus, the learning signal drives $\mX_i$ to incorporate directions it lacks but that are represented by $\mX_j$, encouraging alignment without redundancy. If subspace $j$ is already contained within subspace $i$ i.e., $\tilde{\mP}_j \leq \tilde{\mP}_i$, the gradient vanishes since $\tilde{\mP}_j\tilde{\mP}_i=\tilde{\mP}_j$ implies $(\mI_d-\tilde{\mP}_i)\tilde{\mP}_j = 0$. This update mechanism shares similarities with Oja's rule in online PCA, promoting efficient subspace adaptation.

Conversely, negative pairs induce repulsive gradients, driving $\mX_i$ to remove directions aligned with $\mX_j$ and thus promoting subspace separation. Consequently, the effective dimensionality of subspace $i$ naturally adapts to encompass the union of all its relevant positive neighbors.

\section{WordNet Experiments}
\label{app:sec:wordnet_experiments}
\textsc{WordNet}'s noun hierarchy has 82,115 nodes and 75,850 edges. The verb hierarchy is smaller, featuring 13,767 nodes and 13,239 edges. Their transitive closures are significantly denser, with 663,508 (noun) and 35,079 (verb) edges. All \textsc{WordNet} experiments were conducted on a RTX8000 GPU with 49GB of memory.

\subsection{Reconstruction}
\label{app:sec:wordnet_reconstruction}

\paragraph{Experimental details.} We parameterize each node's subspace with a matrix $\mX_i\in\mathbb{R}^{64\times 64}$ for $\mathrm{SE}^{64}$ and $\mX_i\in\mathbb{R}^{128\times 128}$ for $\mathrm{SE}^{128}$, initialized with entries from a zero-mean Gaussian distribution with standard deviation $0.0001$. The regularizer was set $\lambda=0.2$. For each training edge $(u,v)$, we sample 19 nodes $v' \neq u$ such that neither $(u,v')$ nor $(v',u)$ are in the train split and optimized InfoNCE, applying the the subspace similarity $\mathrm{Tr}(\tilde{\mP}_i\tilde{\mP}_j)$ from Eq.~(\ref{eq:similarity}) to soft projectors. We used Adam \cite{kingma2017adammethodstochasticoptimization}, with a batch-size of 128 and learning rate of $0.0005$. During evaluation, we compute the similarity $\mathrm{Tr}(\tilde{\mP}_u\tilde{\mP}_v)$ of each edge $(u,v)$ in the full transitive closure $\mathrm{TC}(\mathcal{G})$ and rank it among the those of all node pairs that are not connected in the transitive closure $\{\mathrm{Tr}(\tilde{\mP}_u\tilde{\mP}_{v'}) : (u,v')\not\in\mathrm{TC}(\mathcal{G})\}$.

    \label{fig:wordnet_eigenvalues}

\subsection{Link Prediction}
\label{app:sec:wordnet_link_prediction}
\paragraph{Experimental details.} For link prediction, every node is initialized as a random matrix $\mX_i^{d\times n}$, with entries sampled from a zero-mean Gaussian distribution ($\sigma=0.0001$). In our experiments we considered $d=n=64$ as well as $d=n=128$. The soft projector regularizer was set to $\lambda=0.2$. We optimized the margin loss from Eq.~(\ref{eq:margin_loss}) with $\gamma_+=0.9$ and $\gamma_-=0.5$ for 0\% of non-basic edges, and $\gamma_+=0.8$, $\gamma_-=0.1$ for the remaining percentages. To compute this loss, we used 10 negatives per each observed positive edge $(u,v)$. Negatives were generated by sampling 5 corrupted-tail $(u,v')$ and 5 corrupted-head $(u',v)$ examples per positive edge, with corrupted nodes sampled from the entire set of nodes. The results for SE were averaged over 5 random seeds, employing Adam \citep{kingma2017adammethodstochasticoptimization} with a constant learning rate of 0.0005 and a batch-size of 128.

\subsection{Graded Lexical Entailment}
\label{app:sec:hyperlex}
For the \textsc{HyperLex} experiment, we use the noun subset (2,163 pairs), which provides human-annotated scores (0-10) for word pairs $(u,v)$, quantifying the degree to which $u$ is a type of $v$. We quantify entailment using the NIS from Eq. (\ref{eq:nis}), with word sense disambiguation performed as in \citet{athiwaratkun2018hierarchical}, by selecting the \textsc{WordNet} noun synset pair with maximal subspace similarity $\mathrm{Tr}(\tilde{\mP}_i, \tilde{\mP}_j)$. We report the Spearman's rank correlation.

\section{NLI Experiments}
\label{app:sec:nli_experiments}

All NLI experiments use a maximum sequence length of 35 tokens. We train
\texttt{all-MiniLM-L6-v2} and \texttt{all-mpnet-base-v2} with a batch size
of 1024, fine-tuning the transformer backbone end-to-end together with each
model's classification head. Optimization uses Adam
\citep{kingma2017adammethodstochasticoptimization} with a learning rate of
$10^{-4}$, no weight decay, and an exponential per-epoch scheduler with
$\gamma = 0.9$. Label smoothing of 0.1 is applied across all training runs.
Each result is averaged over 5 seeds on a single RTX 8000 GPU (49\,GB).

\paragraph{Subspace embeddings.} The Subspace Projection Head uses 8
attention heads in the MHA pooling layer and $n = d$ learnable query
vectors ($n = 64$ or $n = 128$, matching the ambient dimension). The Beta
priors are initialized as $(\alpha_C, \beta_C) = (1, 6)$ and
$(\alpha_E, \beta_E) = (6, 1)$ and optimized during training. We minimize
cross-entropy on the class posteriors defined in
Eqs.~(\ref{eq:neutral_probability})--(\ref{eq:non_neutral_probability}).

\paragraph{MLP baselines.} Premise and hypothesis embeddings are computed by
mean-pooling the transformer's last hidden state, then passed to a 3-layer
MLP with LeakyReLU activations and hidden dimension matching the transformer
(384 for MiniLM, 768 for mpnet). The final layer outputs 2 or 3 logits
depending on the regime. The MLP and transformer are trained jointly under
standard cross-entropy.

\paragraph{Box embedding baselines.} Mean-pooled transformer outputs are
passed to a 2-layer MLP with LeakyReLU activation and output dimension $2d$,
parameterizing each sentence as a box with corner
$\mathbf{c} \in \mathbb{R}^d$ and positive offset
$\boldsymbol{\delta} \in \mathbb{R}^d_{>0}$, yielding the axis-aligned box
$[\mathbf{c}, \mathbf{c} + \boldsymbol{\delta}]$. Entailment is scored by
the ratio of box-intersection volume to premise volume (the volumetric
analog of the NIS). Since boxes are evaluated only in the 2-way regime, we
omit the neutrality MLP and use the same Beta-posterior framework reduced
to two classes (entail vs.\ non-entail), optimized with cross-entropy.

\subsection{Sensitivity to Hyperparameter $\lambda$}
Table~\ref{app:tab:lambda_analysis} reports SNLI classification accuracy for the all-miniLM-L6-v2 + SPH (SE$^{128}$) model for several values of the regularization parameter~$\lambda > 0$. Across the tested range, model accuracy varies only marginally.

\begin{table}[]
    \caption{\textbf{SNLI test accuracy} for all-miniLM-L6-v2 + SPH (SE$^{128}$) across different values of the hyperparameter~$\lambda$.}
    \centering
    \small
    \begin{tabular}{lcccc}
        \toprule
        & \multicolumn{4}{c}{$\lambda$} \\
         \cmidrule(lr){2-5}
                       & 0.01 & 0.05 & 0.1 & 0.2 \\
        \midrule
         \textbf{2-way}  &  91.18 & 91.26 & 91.12 & 91.06 \\
        \textbf{3-way}   &  85.27 & 85.34 & 85.61 & 85.62 \\
        \bottomrule
    \end{tabular}
    \label{app:tab:lambda_analysis}
\end{table}

\subsection{Compositional Entailment}
\label{app:sec:composite_entailment}

In this section, we describe the construction of the compositional entailment dataset. 

\paragraph{Elicitation.} We prompted Gemini 3.1 Pro to produce SNLI-style
premises and atomic hypotheses. For each premise $p$, the model was asked to
produce two atomic hypotheses ($h_1, h_2$) entailed by $p$ and a third
hypothesis ($h_3$) sharing context with $p$ but factually incompatible with
it. The prompt was:

\begin{quote}\ttfamily\small\raggedright
Generate one example for a natural-language entailment dataset
in the style of SNLI. Each example consists of a premise p describing
a scene or situation, two hypotheses h1 and h2 entailed by p, and a
hypothesis h3 that shares context with p but is factually incompatible.
Output format: <premise>, <hypothesis 1>, <hypothesis 2>, <hypothesis 3>
\end{quote}

\paragraph{Manual validation.} We elicited and reviewed approximately $200$
candidates. An example was discarded if (i) the entailment label of any atomic
hypothesis was judged incorrect, (ii) $h_3$ did not contradict $p$ or did not
share lexical or thematic context, or (iii) any sentence was
ungrammatical or referentially ambiguous. 

\paragraph{Baseline filtering.} We further excluded any examples for which a
baseline misclassified the underlying atomic hypotheses, isolating
composite-level failures from atomic-level errors.

\paragraph{Composite construction.} From each validated premise we form four
composite hypotheses:
\begin{itemize}
    \item \textbf{Entailed:} $h_1 \land h_2$ and $h_1 \land \neg h_3$.
    \item \textbf{Contradicted:} $h_1 \land h_3$ (since $h_3$ contradicts
    $p$) and $h_1 \land \neg h_2$ (since $\neg h_2$ contradicts $p$).
\end{itemize}
 This yielded 600 premise--composite pairs in total: 300 conjunction pairs and 300 negated-conjunction pairs.
 
\paragraph{Example.} For the premise ``Two children are sitting on a red
picnic blanket, eating sandwiches'', with $h_1 = $ ``People are eating'',
$h_2 = $ ``People sitting on a blanket'', and $h_3 = $ ``People sitting
directly on the grass'':
\begin{itemize}
    \item \textbf{Entailed:} $h_1 \land h_2$, $h_1 \land \neg h_3$.
    \item \textbf{Contradicted:} $h_1 \land h_3$, $h_1 \land \neg h_2$.
\end{itemize}

\end{document}